\documentclass{tlp}
\usepackage{natbib}
\usepackage{mathtools}
\usepackage{graphicx}
\graphicspath{ {./images/} }

\def\citep#1{(\citeauthor{#1},~\citeyear{#1})}
\def\citeay#1{\citeauthor{#1},~\citeyear{#1}}
\newcommand{\citeayy}[2]{(\citeay{#1}; \citeay{#2})}
\newcommand{\citeayyy}[3]{(\citeay{#1}; \citeay{#2}; \citeay{#3})}

\usepackage{enumitem}
\usepackage{times}
\usepackage{helvet}
\usepackage{courier}
\usepackage{mathrsfs}
\usepackage[mathscr]{euscript}
\usepackage[all]{xy}
\usepackage{xparse}
\usepackage{verbatim}

\usepackage{csquotes}
\usepackage{lipsum}
\usepackage{listings}

\usepackage{color}
\usepackage{amsmath,amssymb}
\usepackage{latexsym}

\usepackage{xspace}
\usepackage{url}
\usepackage{changepage}
\usepackage[colorlinks=true, allcolors=blue]{hyperref}
\usepackage{float}
\usepackage{wrapfig}
\usepackage{multirow}
\usepackage{hhline}
\usepackage{todonotes}

\usepackage{xspace}
\usepackage{comment}
\usepackage{tikz}
\usetikzlibrary{decorations.markings}

\usetikzlibrary{shapes,arrows}
\tikzstyle{vertexblue}=[circle, fill=blue,draw, inner sep=0pt, minimum size=6pt]
\tikzstyle{vertex}=[circle, draw, inner sep=0pt, minimum size=6pt]
\newcommand{\vertex}{\node[vertex]}

\newcommand{\vertexblue}{\node[vertexblue]}

\tikzstyle{decision} = [diamond, draw, fill=blue!20, 
text width=4.5em, text badly centered, node distance=3cm, inner sep=0pt]
\tikzstyle{block} = [rectangle, draw, fill=blue!20, 
text width=5em, text centered, rounded corners, minimum height=4em]
\tikzstyle{line} = [draw, -latex']
\tikzstyle{cloud} = [draw, ellipse,fill=red!20, node distance=3cm,
minimum height=2em]

\newtheorem{definition}{Definition}
\newtheorem{example}{Example}
\newtheorem{proposition}{Proposition}

\def\wr{:\sim}
 \def\beq{\begin{equation}}
 \def\eeq#1{\label{#1}\end{equation}}
\def\cmodelsdiff{{\sc cmodels(diff)}\xspace}
\def\clingo{{\sc clingo}\xspace}
\def\clingov5{{\sc clingo~5}\xspace}
\def\clingcon{{\sc clingcon}\xspace}
\def\clingolp{{\sc clingo[LP]}\xspace}
\def\clingodl{{\sc clingo[DL]}\xspace}
\def\ezcsp{{\sc ezcsp}\xspace}
\def\ezsmt{{\sc ezsmt}\xspace}
\def\ezsmtPlus{{\sc ezsmt+}\xspace}
\def\ezsmtv3{{\sc ezsmtv3}\xspace}
\def\eezsmtv3{$\mathcal{E}$}
\def\clingconv5{{\sc clingconv5}\xspace}
\def\eclingcon{{\sc c-con}\xspace}
\def\eclingodl{{\sc c[DL]}\xspace}
\def\cP{\ensuremath{\mathcal{P}}}
\def\cvcFour{{\sc cvc4}\xspace}
\def\cvcFive{{\sc cvc5}\xspace}
\def\zThree{{\sc z3}\xspace}
\def\yices{{\sc yices}\xspace}
\def\gringo{{\sc gringo}\xspace}
\def\level#1{{\lambda({#1})}}

\newcommand{\weak}[2]{\text{weak}(#1,#2)}
\newcommand{\pasw}[3]{#1_{#2}^{#3}}

\newcommand{\At}{\mathit{At}}

\def\true{\mathit{true}}
\def\false{\mathit{false}}
\def\cB{\ensuremath{\mathcal{B}}}

\def\cF{\ensuremath{\mathcal{F}}}

\def\cP{\ensuremath{\mathcal{P}}}

\def\true{\ensuremath{\mathit{true}}}
\def\false{\ensuremath{\mathit{false}}}

\def\ez{{CAS}\xspace}

\def\ar{\leftarrow}
\def\rar{\rightarrow}

\begin{document}

\jnlPage{1}{8}
\jnlDoiYr{2021}
\doival{10.1017/xxxxx}

\title[EZSMT Version 3]{EZSMT Version 3, Matured
\thanks{Partially funded by UNO GRACA 2024}}

\begin{authgrp}
\author{\sn{Keeran} \gn{Dhakal}}
\affiliation{University of Nebraska Omaha, USA}
\author{\sn{Yuliya} \gn{Lierler}}
\affiliation{University of Nebraska Omaha, USA}
\end{authgrp}


\maketitle              
\begin{abstract}
Constraint Answer Set Programming (CASP) is a hybrid reasoning paradigm that combines Answer Set Programming (ASP) with Constraint Processing and Satisfiability Modulo Theories (SMT), enabling powerful declarative encodings of complex combinatorial search problems. 
This paper presents the design and implementation of \ezsmtv3, an extensible SMT-based CASP framework that advances the translational approach to CASP solving. Building upon the foundation of the \ezsmtPlus system, \ezsmtv3 introduces a more expressive input language, supports optimization via weak constraints, and offers foundations for streamlined  integration of new constraint types. 
Rather than implementing custom search procedures, \ezsmtv3 leverages state-of-the-art SMT solvers, such as \cvcFive, \yices, and \zThree  to perform reasoning. 
The paper provides benchmarking results comparing  \ezsmtv3 with its CASP peers such as \clingcon, \clingodl, and \clingolp, while showcasing its ability to handle mixed-domain constraints involving both integers and reals. 
The system provides a robust platform for future extensions and theoretical exploration within the CASP domain. Under consideration in Theory and Practice of Logic Programming (TPLP).
\end{abstract}

\section{Introduction}
Constraint answer set programming (CASP) is a hybrid methodology in automated reasoning that integrates advancements from several research domains, namely answer set programming~(\citeay{nie99}; \citeay{mar99}; \citeay{bre11}), constraint processing~\citeayy{ros08}{jm94}, and satisfiability modulo theories~\citeayyy{nie06}{BarretSST08}{BarTin-14}. Works by~\citeayyy{elk04}{mel08}{lier14} are among earlier references to CASP. It has shown significant potential, leading to the creation of numerous solvers such as 
 {\sc acsolver}~\citep{mel08},
{\clingcon}~\citep{geb09}, {\ezcsp}~\citep{lierbal17}, 
{\sc idp}~\citep{idp}, {\sc inca}~\citep{dre11a}, {\sc
  dingo}~\citep{jan11},  {\sc mingo}~\citep{liu12}, {\sc aspmt}~\citep{Bartholomew2014}, \clingolp and \clingodl~\citep{jan17}, and~\ezsmtPlus~\citeayy{sus16b}{shen18ezsmt}.
  CASP opens up new possibilities for declarative programming, enabling it to tackle such complex tasks as train scheduling and product configurations. Solvers for CASP can be broadly categorized based on their construction strategy into integrational and translational approaches. This paper describes not just a solver that practices translational approach but an extensible CASP framework that is geared to ease the implementation of new systems in this field.

This paper presents the design, development, and implementation of an extensible SMT-based constraint answer set programming framework  \ezsmt version 3 (\ezsmtv3). We build upon the initial vision outlined in our earlier work on the \ezsmtPlus system~\citeayy{sus16b}{shen18ezsmt}. In particular, we continue championing the practice of so called translational approaches within the automated reasoning realm.
The work on \ezsmtv3 turns preliminary ideas behind the CASP  \ezsmtPlus solver into mature extensible framework for CASP. With that, not only \ezsmtv3 is the solver itself, it is also designed to support  extensions of this system to new kinds of constraints in a simple, streamlined manner.
In a nutshell,
the \ezsmtv3 system computes answer sets to constraint answer set (CAS) programs providing support for various kinds of constraints. Yet, while doing so it does not implement native search procedures. Instead, it translates a given logic program with constraint atoms into a formula within some dialect of satisfiability modulo theories (SMT). This formula is then processed by one of the off-the-shelf SMT solvers.  
Historically, the \ezsmtPlus language adopted the conventions of the CASP language developed for the {\sc ezscp} system~\citep{lierbal17}. 
Thus, its constraint atoms  (marked by the keyword {\tt required}) were restricted to rule heads, making certain domains cumbersome to formalize. While sufficient for bootstrapping a proof-of-concept system, {\ezcsp}’s language features  revealed the need for a more expressive and flexible alternative. 

In our work on \ezsmtv3, we found such an alternative.
 In particular, it builds upon the  developments in the \clingo~5 series~\citeayy{multishot19}{kaminski2023build} that promotes  the extensibility philosophy. The \clingov5 system provides means to 
elaborate the specifications for new kinds of constructs to be incorporated for processing within its grounding tool  \gringo~\citeayyy{gekaosscth09a}{geb15}{kaminskiPhD}. 
In addition,  \clingov5 provides means to incorporate custom propagators to ensure proper processing of newly incorporated syntactic language features. 
In this work, we embrace the extensibility philosophy of \clingov5. Yet, we diverged from its provisions for the custom implementations of the search mechanisms. We advocate the utilization of already existing state-of-the-art automated reasoning tools, specifically, SMT solvers. Thus, this work relies on a body of theoretical findings relating CASP and SMT as well as  a body of sophisticated algorithmic developments within SMT solving resulting in such exemplary systems as \cvcFour  \citep{barrett2011cvc4}, \cvcFive \citep{barrettcvc5}, \zThree \citep{de2008z3}, and \yices \citep{dutertre2006yices}.
Instead, we focused on creating a streamlined interface to SMT technologies. 
 Our approach relies on an easily extensible component-based architecture consisting of a grounding component, a translation component, and an SMT-based solving component. The system is designed as a multi-stage processing pipeline, with the output of each component serving as the input for the next. The grounding component is utilized in such a way that the system allows us to extend the existing language specifications. The output of the grounding component is interpreted by the translation component to support translations from ASP to SMT—paving the way for future CASP dialects to be seamlessly integrated. The SMT-based solving component uses the translated SMT-LIB program to return the answer sets of the original ASP program.
 In addition, \ezsmtv3 implements support for optimization statements, namely, weak constraints. This feature is new to \ezsmtv3 and was missing from the  CASP  \ezsmtPlus solver. 


Section~\ref{sec:background} of this  paper  provides a  review of key concepts in constraint answer set programming. Section~\ref{sec:ezsmt_all} starts by detailing the CASP dialects supported by \ezsmtv3 by utilizing a formalization of  a variant of the Traveling Salesman problem as our running example. It concludes with the 
 presentation on the architecture of the \ezsmtv3 system and the discussion of  its implementation. Given that optimization statements are new to \ezsmtv3 in relation to its older ``sibling'' \ezsmtPlus, Section~\ref{sec:optimizations} introduces syntax and semantics of language constructs used to express optimization statements within programs supported by the system. This section concludes with the details on the implementation.
In Section~\ref{sec:bench}, we discuss results on  benchmarking the performance of \ezsmtv3 against its closest CASP relatives such as \clingcon, \clingolp and \clingodl. We note that the capabilities of \ezsmtv3 extends beyond any of these peers as, for example, the system is capable to support reasoning with constraint atoms that contain both integer and real variables. 
At last we remark on future work.

\section{Background}\label{sec:background}
\subsection{Logic Programs and Input Answer Sets}
Many definitions presented in this section follow the lines by~\citeauthor{lier23}~\citeyearpar[Sections 3 and 4]{lier23}.
\paragraph{Logic programs}
A {\em vocabulary} is a set of propositional symbols, also called atoms.
A {\em literal} is an atom $a$ or its negation $\neg a$.
A \emph{(propositional) logic program} over vocabulary~$\sigma$  is a
set of \emph{rules} of the form
%
\begin{equation}\label{e:rule}
\begin{array}{l}
a\ar b_1,\ldots, b_\ell,\ not\  b_{\ell+1},\ldots,\ not\  b_m,\ 
\ not\  \ not\  b_{m+1},\ldots,\ not\  \ not\  b_n.
\end{array}
\end{equation}
where $a$ is an atom in $\sigma$ or $\bot$, and each $b_i$, where $1\leq i\leq n$,
is an atom in $\sigma$.
We will  use the abbreviated form of a rule~\eqref{e:rule}, i.e.,
\begin{equation}\label{e:ruleabr}
a\ar B.
\end{equation}
where $B$ stands for the right hand side of the arrow in~\eqref{e:rule}, 
and is also called a {\em body}. By $B^+$ we denote the {\em positive} part of body $B$, i.e., $b_1,\ldots, b_\ell$. We sometimes identify body $B$
with the propositional formula
\begin{equation}\label{e:body-formula}
b_1\wedge\ldots\wedge b_\ell\wedge \neg  b_{\ell+1} \wedge\ldots\wedge\neg  b_m \wedge \neg\neg b_{m+1} \wedge\ldots\wedge \neg\neg  b_n.
\end{equation}
and rule~\eqref{e:rule} with the propositional formula (implication)~\hbox{$B\rar a$}.
The expression $a$ is the \emph{head} of the rule.  A rule whose head is the symbol $\bot$ is called a {\em denial}.
 A rule~\eqref{e:ruleabr} whose body is empty, i.e., $n=0$ is called a {\em fact}; in this case it is frequently written as $a.$ (while $B$ is identified with $\top$ and $\top\rar a$ is identified with $a$).
For a logic program $\Pi$ (a propositional formula~$F$),
by $\At(\Pi)$ (by $\At(F)$) we denote the set of atoms occurring in $\Pi$ (in~$F$).

It is customary for a given vocabulary $\sigma$, to identify a set $X$ of atoms over $\sigma$ with (i) a complete and consistent set of literals over $\sigma$ constructed as $X\cup\{\neg a \mid a\in\sigma\setminus X\}$, and respectively with (ii)~an assignment function or interpretation that assigns truth value $\true$ to every atom in~$X$ and~$\false$ to every atom in $\sigma\setminus X$. 
Within the scope of this paper, we are interested in sets of atoms in relation to respective programs, so that the signature of that program will be considered for reference.
We say a set~$X$ of atoms {\em satisfies} rule~\eqref{e:ruleabr}, if~$X$ (understood as an assignment function) satisfies the propositional formula $B\rightarrow a$.
Naturally, we can speak about set $X$ satisfying the body or the negative part of the body of the rule as we identify these with respective propositional formulas. 
 We say~$X$ satisfies a program~$\Pi$, if~$X$ satisfies every rule in~$\Pi$. In this case, we also say that $X$ is a model of $\Pi$. We may abbreviate the satisfaction relation with symbol $\models$ (to denote that a set of atoms satisfies a rule, a body,  a program, or a formula).

The {\sl reduct} $\Pi^X$ of a program $\Pi$ relative to a set $X$ of atoms is 
obtained by first removing all rules~\eqref{e:rule} such that $X$ does not satisfy the negative part of the body
$$\neg  b_{\ell+1} \wedge\ldots\wedge\neg  b_m \wedge \neg\neg b_{m+1} \wedge\ldots\wedge \neg\neg  b_n,$$ 
and replacing all remaining rules with~$a\ar b_1,\ldots, b_\ell$. 
\begin{definition}[Answer set]
	\label{def:answer-set}
A set~$X$ of atoms is an {\em answer~set}, if it is the minimal set that satisfies all rules of $\Pi^X$~\citep{lif99d}. 
\end{definition}

\begin{example}
Consider a program 
\beq
\begin{array}{l}
b\ar a.\\
c\ar not\ a.\\
\end{array}
\eeq{eq:p1input}
This program has a single answer set, namely, $\{c\}$.
Let us construct a new program from program~\eqref{eq:p1input} by appending a single fact to it.
\beq
\begin{array}{l}
a.\\
b\ar a.\\
c\ar not\ a.\\
\end{array}
\eeq{eq:p1nc}
This program has a single answer set, namely, $\{a,b\}$. 
\end{example}

Consider now another program
\beq
\begin{array}{l}
a\ar not\ not\ a.\\
b\ar a.\\
c\ar not\ a.\\
\end{array}
\eeq{eq:p2c}
The first rule of this program is typically written as $$\{a\}.$$ 
Rules of this form are called  {\em choice rules}.
We can intuitively read the rule above as ``{\em atom $a$ may be the case}''. This program has two answer sets: 
\beq
\hbox{$\{a,b\}$ and $\{c\}$. }
\eeq{eq:as1.1}

We now state the definition of an input answer set, 
as it is instrumental in defining constraint answer set programs.
\begin{definition}[Input Answer Set]
	\label{def:input-answer-set}
	For a logic program $\Pi$ over vocabulary~$\sigma$ and vocabulary $\iota\subseteq\sigma$ such that none of $\iota$'s elements occur in the heads of rules in $\Pi$,
	a set~$X$ of atoms over~$\sigma$ is an \emph{input answer set} of~$\Pi$ relative to  $\iota$,  when $X$ is an answer set of the program
	$\Pi\cup (X\cap\iota)$.
\end{definition}
Recall program~\eqref{eq:p1input},
which has two input answer sets relative to signature $\{a\}$. These answer sets are listed in~\eqref{eq:as1.1}. 

The reader may obtain new insights about the definition of an input answer set in light of the following formal result.
\begin{proposition}
	For a logic program $\Pi$ over vocabulary~$\sigma$ and vocabulary $\iota\subseteq\sigma$ such that none of $\iota$'s elements occur in the heads of rules in $\Pi$, the answer sets of a program $$\Pi\cup \{~\{a\}.~\mid a\in \iota\}$$ coincide with the input answer sets of program $\Pi$ relative to $\iota$.
\end{proposition}
Thus it is not by chance that answer sets of program~\eqref{eq:p2c} and input answer sets of program \eqref{eq:p1input} relative to signature $\{a\}$ coincide.

The {\em dependency graph} of~$\Pi$ is the directed graph $G$ such that 
\begin{itemize}
\item the vertices of $G$ are the atoms occurring in~$\Pi$, and
\item for every rule~\eqref{e:rule} in~$\Pi$ whose head is not $\bot$, $G$ has an edge from atom $a$ to each atom in positive part $b_1,\dots, b_\ell$ of its body.
\end{itemize}

A program is called {\em tight} if its dependency graph is acyclic. It is easy to see that any sample program discussed so far is tight. The simplest nontight program is as follows $a\ar a.$

\subsection{Constraints, CSP, SMT}
Lierler and Susman~\citeyearpar{lie17} illustrated that the notion of a  constraint syntactically coincides with ground literals of satisfiability modulo theories (SMT). Furthermore, a  constraint satisfaction problem (CSP) --- posed as a set of constraints --- can be identified with a conjunction of ground literals, which is evaluated by means of first-order logic interpretations/structures  representative of a particular ``uniform" SMT-logic~\citep{lie17}. Thus, in a way we can understand satisfiability modulo theories via the lens of satisfiability modulo constraints. 

Intuitively, uniform SMT-logics are defined via interpretations/structures whose domain, interpretation of ``theory/constraint/interpreted'' predicate symbols, and ``interpreted" function symbols are fixed.
In practice, special forms of  constraints are commonly used. {\em Integer linear constraints} are examples of these special cases. 
Let us recall their syntactic shape and provide some intuitions for their interpretations prior to diving into formal settings.
An {\em (integer) linear expression} 
has the form
\begin{equation}
a_1 x_1 +\cdots + a_n x_n, 
\label{eq:exp}
\end{equation}
where $a_1,\dots,a_n$ are (integer) numbers and $x_1,\dots,x_n$ are  (constraint) variables whose domain ranges over (integer) numbers. Note how this definition encapsulates both integer linear expressions and  linear expressions over real numbers. For the latter,  we drop the word integer as a requirement on coefficients and variables. 
It is customary to omit  coefficients when their value is~$1$ and also replace $+$ by $-$ when the coefficient is a negative number, while that negative number is replaced by its absolute value. The SMT technology utilizes the standard SMT-LIB language~\citep{BarST-SMTLIB}.  In that language prefix notation is used so that 
expression~\eqref{eq:exp} is written as
\[
+(\times(a_1,x_1),+(\times(a_2,x_2),\dots +(\times(a_{n-1},x_{n-1}),\times(a_n,x_n))\dots).
\]
We call a constraint  {\em (integer) linear} when it has the form 
\begin{equation}
e\bowtie k
\label{eq:lc}
\end{equation}
where $e$ is  (an integer) linear expression, $k$ is (an integer) number, and $\bowtie$ belongs to 
\begin{equation}
\{<,>,\leq,\geq,=,\neq\}.  
\label{eq:arithmop}
\end{equation}
We can write~\eqref{eq:lc} as an expression $\bowtie(e,k)$ in prefix notation. When $e$ is also written in prefix notation, it is easy to see how this constraint takes the shape of a ground atom. Let us agree to call these kinds of atoms {\em constraint (ground) atoms}. In the sequel, we use the terms {\em constraint} and {\em constraint atom} interchangeably.


For instance, consider   an integer  linear  constraint 
\begin{equation}
    2 x+ 3 y>0.
\label{eq:lia}\end{equation} 
When written in prefix notation it takes the shape of constraint ground atom
$$>(+(\times(2,x),\times(3,y)),0),$$
where
\begin{itemize}
    \item $>$ is a binary ``interpreted'' predicate symbol; 
\item $+$ and $\times$ are binary ``interpreted'' function symbols;
\item $0$, $2$, and $3$ are 0-arity ``interpreted'' function symbols; and 
\item $x$ and $y$ are 0-arity ``un-interpreted'' function symbols.
\end{itemize}
In the logic literature, 0-arity un-interpreted function symbols
are frequently referred to as  object constants, whereas in the constraint processing literature they are referred to as (constraint) variables. Here, we use the term {\em constraint variables}.
Prior to some formal definitions, let us talk about this constraint atom as a formula within satisfiability modulo  Linear Integer Arithmetic Logic, intuitively.
This logic is defined by interpretations, such that
\begin{itemize}
\item 
$>$ is interpreted as an arithmetic greater than relation; 
\item 
$+$ and $\times$ are interpreted as usual in arithmetic; 
\item 
$0$, $2$, and $3$ are mapped into respective integer domain elements identified with the same symbol, thus we may also refer to these function symbols as integers; and  
\item 
$x$ and $y$ are mapped into integers; in general,   the  domain for 0-arity function symbols occurring in constraint  atoms is the set of integers.
\end{itemize}

In naming the constraints, we use conventions adopted by SMT-LIB\footnote{\url{https://smt-lib.org/logics.shtml}} so that
\begin{itemize}
\item IA stands for the theory Ints (Integer Arithmetic);
\item RA stands  for the theory Reals (Real Arithmetic);
\item IRA stands for the theory Reals and Ints (mixed Integer Real Arithmetic);
\item IDL stands for Integer Difference Logic;
\item L before IA, RA, or IRA stands for the linear fragment of those arithmetics.
\end{itemize}
Let us call 
\begin{itemize}
\item integer linear constraints --- {\em LIA constraint atoms (LIA constraints)};
\item linear constraints --- {\em LRA constraint atoms (LRA constraints)};
\item constraints that syntactically have the form of expression~\eqref{eq:lc}, while the coefficients and constraint variables of this expression can be both integer and real numbers
--- {\em mixed integer real constraint atoms (LIRA constraints)};
\item constraints that have the form
\begin{equation}
x-y\bowtie c \hbox{~~~~~~    or~~~~~~ }
x\bowtie y,
\label{eq:idlcon}    
\end{equation}
where $\bowtie$ is one of the arithmetic relations in~\eqref{eq:arithmop},
$x$ and $y$  are constraint variables over integers, and $c$ is an integer --- {\em IDL constraint atoms (IDL constraints)}.
\end{itemize}

We are now ready to provide formal definitions for four  logics within SMT framework utilized in this work.
\begin{definition}[Satisfiability Modulo Theories Formula or SMT Formula]
{\em Formula in satisfiability modulo  Linear Integer Arithmetic Logic  or
 SMT(LIA) Formula}  ---
is a variable-free first order logic formula that consists of propositional  or  LIA constraint atoms.
Its {\em interpretations} can be captured by {\em valuations} -- functions -- that map all propositional atoms to truth values and constraint variables to integers; while arithmetic predicate and function symbols are interpreted as customary in arithmetic. 

{\em SMT(LRA) formulas or SMT formulas in  LRA  Logic},
{\em SMT(LIRA) formulas or SMT formulas in  LIRA  Logic},
{\em SMT(IDL) formulas or SMT formulas in  IDL  Logic}  are defined similarly considering LRA, LIRA, and IDL constraint atoms, respectively, in place of LIA constraint atoms. 
Interpretations for these formulas are  captured by valuations that respect domains of the constraint atoms according  to their types.

{\em Models} of SMT formulas are interpretations that satisfy SMT formulas, where the satisfaction relation is understood classically as in first order logic. 
\end{definition}

\begin{example}
For instance, the SMT(LIA) formula
\begin{equation}
p\rightarrow\big((x\geq 1 \wedge x \leq 3)\vee x=5\big).
    \label{eq:ex:smlia1.0}
\end{equation}
consists of propositional atom $p$ and LIA constraint atoms 
$$
\begin{array}{lll}
x\geq 1 ~~~~~~~& x \leq 3 ~~~~~~~& x=5.
\end{array}
$$
This formula has four models when $p$ is interpreted as $\true$ captured by the following valuations 
$$
\begin{array}{ll}
x\mapsto 1 ~~~~~~~& p\mapsto \true\\ 
x\mapsto 2 ~~~~~~~& p\mapsto \true\\
x\mapsto 3 ~~~~~~~& p\mapsto \true\\
x\mapsto 5 ~~~~~~~& p\mapsto \true\\
\end{array}
$$
There are an infinite number of models for this formula  when $p$ is interpreted as $\false$ including, for example, one captured by the  valuation  
$$
\begin{array}{ll}
x\mapsto 4 ~~~~~~~& p \mapsto \false.
\end{array}
$$
\end{example}
We are now ready to state definitions for constraint satisfaction problems.

\begin{definition}[Constraint satisfaction problem or CSP]
We call a finite set of constraints a
{\em constraint satisfaction problem (CSP)}. Within this work we will consider CSPs of particular kind. Namely, 
\begin{itemize}
\item {\em Linear Integer Arithmetic CSP (LIA CSP)}
 formed as a set of LIA constraints;   
\item {\em Linear Real Arithmetic CSP (LRA CSP)} 
     formed as a  set of LRA constraints;   
\item {\em LIRA CSP} formed as a  set of LIRA constraints;
\item {\em Integer Difference Logic CSP (IDL CSP)} formed as a finite set of IDL constraints.
\end{itemize}
As we  identify  constraints with ground atoms, we also identify a CSP with a conjunction of constraints/ground atoms occurring in its set. Thus, any LIA CSP, LRA CSP, LIRA CSP, and IDL CSP can be viewed as a special form of SMT(LIA), SMT(LRA), SMT(LIRA), SMT(IDL) formula, respectively.
We call models of these formulas   {\em solutions} of respective CSPs. 
\end{definition}
\begin{example}
One of the solutions to the LIA CSP composed of a single constraint~\eqref{eq:lia}  is a valuation  that maps $x$ to $0$ and $y$ to $1$. If we are to form another LIA CSP -- a set composed of constraint~\eqref{eq:lia} and constraint $y\neq 1$, then  the valuation that maps $x$ to $0$ and $y$ to $1$ is not a solution, while, for instance, a valuation that maps $x$ to $0$ and $y$ to $2$ is.
\end{example}

Both LRA and IDL CSPs are interesting from the perspective that there are tractable algorithms to decide whether 
these problems have solutions.
This is not the case for LIA  and LIRA CSPs.

\subsection{Constraint answer set programs and their relation to SMT}\label{sec:casp}
Let $\sigma_r$ and $\sigma_i$ be two disjoint vocabularies.
We refer to their elements as \emph{regular}  and \emph{irregular}\footnote{In the literature on constraint answer set programming, atoms of this kind are frequently called {\em constraint} atoms. In the literature on satisfiability modulo theories, atoms with similar role are called {\em theory} atoms.} atoms, respectively.
\begin{definition}[Constraint Answer Set Program or CAS Program]
	Let $\sigma=\sigma_r
	\cup\sigma_i$ be a vocabulary so that $\sigma_r$
	and $\sigma_i$ are disjoint; $\cB$ be a set of constraints; $\gamma$ be an injective
	function from the set of irregular literals over~$\sigma_i$ to  $\cB$.

	We call a triple~$P=\langle \Pi,\cB,\gamma\rangle$ a {\em \ez program}
	over vocabulary $\sigma_r
	\cup\sigma_i$, when
	$\Pi$ is a logic
	program over  $\sigma_r
	\cup\sigma_i$ such that
	any rule that contains  atoms in $\sigma_i$ is a rule with symbol $\bot$ in its head.

	A set $X\subseteq\At(\Pi)$ of atoms
	is an \emph{answer set}
	of $P$ if
	\begin{enumerate}
		\item[(a)]  $X$ is an input answer set of $\Pi$ relative to $\sigma_i$, and
		\item[(b)] the  following CSP           has a solution:
		$\{\gamma(a) \mid a\in X\cap\sigma_i\}\cup \{\gamma(\neg a) \mid a\in \sigma_i\setminus X\}.
		$
	\end{enumerate}
	A pair $\langle X,\nu\rangle$
	is an \emph{extended answer set}
	of $P$ if $X$ is an \emph{answer set} of $P$ and valuation~$\nu$ is a solution to the CSP constructed in (b).

Within this work we  consider CAS programs $\langle \Pi,\cB,\gamma\rangle$ of a particular kind. Namely, 
\begin{itemize}
\item {\em CAS(LIA) programs} whose set $\cB$ of constraints is
 formed by LIA constraints;   
\item {\em CAS(LRA) programs}  whose set $\cB$ of constraints is formed by LRA constraints;   
\item {\em CAS(LIRA) programs}  whose set $\cB$ of constraints is formed by LIRA constraints;   
\item {\em CAS(IDL) programs}  whose set $\cB$ of constraints is formed by IDL constraints.
\end{itemize}

\end{definition}
It is due to note that when CAS programs are written in practice, the CASP systems permit a user listing an irregular atom in the head of the rule. Yet, that should be considered as  ``syntactic sugar'' so that a rule of the form~\eqref{e:ruleabr}, where $a$ is an irregular atom is seen as an abbreviation for the rule $$\ar B,\ not\ a.$$
In the presentation, we  utilize vertical bars to mark the irregular atoms which will  have intuitive mappings into their respective  constraints. For instance, irregular atom $|x\geq12|$ naturally maps into constraint $x\geq12$.

\begin{example}\label{example:cas}
We now exemplify the definition of a CAS program. Let $\Pi_1$ be logic program~\eqref{eq:p2c} extended with a denial 
$$\ar a,\ |x\geq12|,$$ 
where $|x\geq12|$ denotes an irregular atoms with constraint variable $x$. Let 
$\cB_1$ be a set of integer linear constraints 
$\{x\geq 12,x< 12\}$; 
$\gamma_1$ be an injective
	function from irregular literals in the signature of $\Pi_1$ to constraints
  $$|x\geq12|\rar x \geq 12,\ \ \ \   
  \neg |x\geq12|\rar x < 12.
  $$  
CAS program 
$\langle \Pi_1,\cB_1,\gamma_1\rangle$ has three answer sets, namely,
$$
\begin{array}{l}
\{a,b\}\\
\{c\}\\
\{c,|x\geq12|\}\\
\end{array}
$$
and infinitely many extended answer sets: 
$$
\begin{array}{lll}
\{a,b,x\mapsto 11\}~~~&\{a,b,x\mapsto 10\}~~~&\{a,b,x\mapsto 9\}\dots\\
\{c,x\mapsto 11\}~~~&\{c,x\mapsto 10\}~~~&\{c,x\mapsto 9\}\dots\\
\{c,|x\geq12|,x\mapsto 12\}~~~ &\{c,|x\geq12|,x\mapsto 13\}~~~&\{c,|x\geq12|,x\mapsto 14\}\dots\\
\end{array}
$$
\end{example}

 We  refer to CAS program  $P=\langle \Pi,\cB,\gamma\rangle$ as {\em tight} when its first member $\Pi$ has this property.

 \citeauthor{lie17}~\citeyearpar{lie17} illustrated that for  CAS programs of the four kinds considered here, one can construct an SMT formula (of the four kinds considered here) so that its models coincide with the extended answer sets of the given program. They generalized  the concepts of completion and level ranking -- originally introduced by \cite{cla78} and \cite{nie08}, respectively -- which are essential in the construction of such an SMT formula.  Intuitively, completion is a process that turns a CAS program into an SMT formula. This formula comes with a special guarantee that every extended answer set of the given program is a model of its completion. For the class of tight programs the reverse direction is also the case. As a result, the extended  answer sets of a CAS program  coincide with the models of its completion.
 In case of a program being nontight, so called level ranking constraints added to a completion will ensure that computed models (modulo newly introduced integer variables within level ranking constraints) are exactly the answer sets.
We now provide  details of that translation relevant to understanding the workings of the \ezsmtv3 system.

Within the translation, irregular atoms are introduced that encode level ranking constraints required to weed out models of the completion that are not answer sets. For instance, an irregular atom
$|lr_a-lr_{b}\geq 1|$  encodes an IDL (or LIA or LIRA) constraint
\hbox{$lr_a-lr_{b}\geq 1$}, where $lr_a$ and $lr_b$ are integer constraint variables.
Let 
\hbox{$P=\langle \Pi,\cB,\gamma\rangle$} be a CAS program over  $\sigma_r\cup\sigma_i$.
If a program is not tight, for every atom $a\in\sigma_r$ that occurs in~$\Pi$, we introduce an integer variable~$lr_a$. The SMT formula $\cF^P$ is constructed as  a conjunction of the following
\begin{enumerate}
	\item implications corresponding to rules~\eqref{e:rule}  in $\Pi$;
	\item
	for each regular atom $a$ occurring within the given CAS program, the implication
 \begin{itemize}
     \item  
	$
	a\rar
	\displaystyle\bigvee_{a\ar B\in \Pi }B$, when the program is tight
 
 \item 	
	$
	a\rar	\displaystyle\bigvee_{a\ar B\in \Pi }\big(B\wedge
	\bigwedge_{b\in B^+\setminus\sigma_i}   
	|lr_a - lr_{b}|\geq 1
	\big)$, otherwise;
  \end{itemize}

 \item	for each irregular atom $|c|\in\sigma_i$ occurring within the given CAS program (where $c$ is a constraint; recall that irregular atoms are assumed to have a natural mapping into respective constraints), the
  equivalence $|c|\ \longleftrightarrow\ c$;
  \item	in case the considered program is not tight, for each irregular atom of the form 
  \hbox{$|lr_a - lr_{b}\geq 1|$} introduced within the translation, the equivalence $$|lr_a-lr_{b}\geq 1|\ \longleftrightarrow\ lr_a-lr_{b}\geq 1.$$
\end{enumerate}
In case of a tight CAS program $P$, formula $\cF^P$ captures the completion of $P$.

\begin{example}\label{example:comp}
Recall CAS program 
$\langle \Pi_1,\cB_1,\gamma_1\rangle$
from Example~\ref{example:cas}. Let us call it $P_1$. $\cF^{P_1}$ is as follows
$$
\begin{array}{llll}
\neg \neg a\rar a~~~~& a\rar b~~~~& \neg a\rar c~~~~&  a\wedge \ |x\geq12|\rar \bot\\
a\rar \neg \neg a&b\rar a&c\rar \neg a& \\
 |x\geq12| \longleftrightarrow x\geq 12&&&\\
\end{array}
$$
The models of this formula coincide with the answer sets of $\langle \Pi_1,\cB_1,\gamma_1\rangle$.
\end{example}

\begin{figure}[th]
\begin{tabular}{l|c|c}
    &~~~~~~~Tight~~~~~~~ & ~~~~~~~Non-Tight~~~~~~~\\
    \hline\hline
    CAS(LIA)&\multicolumn{2}{c}{SMT(LIA)}\\
    \hline
    CAS(LRA)&{~~~~~~~SMT(LRA)~~~~~~~}&{~~~~~~~SMT(LIRA)~~~~~~~}\\
    \hline
  CAS(LIRA)~~~~~~~&\multicolumn{2}{c}{SMT(LIRA)}\\
  \hline
    CAS(IDL)&\multicolumn{2}{c}{SMT(IDL)}\\
    \end{tabular}
    \caption{Mapping of CAS programs to respective SMT formulas.\label{fig:castosmt}}
\end{figure}

Figure~\ref{fig:castosmt} summarizes the details on which kind of SMT formula system \ezsmtv3 obtains during the application of the described translation process depending on the properties of the given CAS program. For instance, row 2 in the table of this figure states that given a CAS(LRA) program which is 
\begin{itemize}
\item tight, the translation  results in SMT(LRA) formula;
\item non-tight, the translation results in SMT(LIRA) formula. \end{itemize}

\section{\ezsmt Version 3 Language(s), Use Case, and Architecture}\label{sec:ezsmt_all}
This section is devoted at large to the description of the language and architecture of the \ezsmt Version 3 system, abbreviated as \ezsmtv3. Prior to providing the details on the system's components, we  articulate a bird's-eye view on the system by pointing at its major design choices. We also provide a sample use case of the system utilizing the Traveling Salesman problem. The presentation of this use case is intermixed with the  details on the syntactic constructs supported by the \ezsmtv3 together with their mappings into respective CAS fragments.

In a way, \ezsmtv3 can be seen as a system that puts together the ideas and practices behind two CASP solvers, namely, \clingcon version 3 (and above)~\citep{ost17} and  \ezsmtPlus \citep{shen18ezsmt}. In particular, from \clingcon it borrows an idea to utilize capabilities unique to the 
grounder \gringo version~5~\citep{gebser2016theory}. 
This grounder provides a possibility
\begin{itemize}
  \item to specify the grammar of the language of constraints of interest and 
\item to use that newly defined language in writing programs that are subsequently grounded by \gringo.  
\end{itemize}
  From \ezsmtPlus, \ezsmtv3 borrows an idea to utilize  an SMT solver, such as \zThree or \cvcFive, as its search engine back-end after computing completion and level rankings of a given CAS program. 
The combination of the \gringo version~5 front-end and an SMT solver as a back-end uniquely positions system \ezsmtv3 not only as a CASP solver but also as an easily extensible framework for creating new kinds of CASP solvers. Indeed, SMT solvers support a multitude of  distinct logics -- languages for specifications of constraint atoms -- while \gringo version~5 allows us to specify a language of such constraint atoms and quickly incorporate these within the grounding stage of processing. 
The major routines of building completion and then translating that internal representation into the standard language supported by SMT solvers, namely, SMT-LIB is something that
\ezsmtv3 inherits from \ezsmtPlus and provides as part of the framework for extensions to new logics.
  In the sequel, we omit the reference to version of 
  \gringo assuming  version 5 as default.

\subsection{\ezsmtv3 Language(s) and  Its Use Case}\label{sec:languc}
We start this section by uncovering the details of the \ezsmtv3 language used for formulating programs in CAS(LIA).
We then present the CAS(LIA) formalization of a variant of the  Traveling Salesman (TS) Problem~\citeayy{tsp1}{tsp2}. The presented CAS(LIA) program is written in the language supported by  systems \ezsmtv3 (and \clingcon). The similar formalization of the TS problem was presented by~\cite{lier23} in the language supported by the CASP solvers {\ezcsp}~\citep{lierbal17} (and  \ezsmtPlus).
At last, we discuss the details of \ezsmtv3 language used for formulating programs in CAS(LRA),  CAS(LIRA), and CAS(IDL).

\subsubsection{\ezsmtv3 CAS(LIA) Language}\label{sec:caslialang}
As mentioned earlier, system \gringo is used within \ezsmtv3 as a front-end to ground a considered CAS program. Section~\ref{sec:tsp} demystifies the process of grounding.  It presents the Traveling Salesman problem encoding that is formalized using CAS(LIA) ``schemata'' rules --- rules that contain ``ASP'' variables and hence can be seen as abbreviations for groups of corresponding ground/propositional CAS(LIA) rules such as presented in preliminaries. 
Within this section, we consider ground/propositional programs for simplicity.

\begin{lstlisting}[
  caption = {Encoding of LIA Logic in \gringo version 5. 
%   The program is annotated with \emph{achievements}.
},
  label={list:liatheory},
  basicstyle=\ttfamily\small,
  numbers=left,
  stepnumber=2,
]
#theory lia {
    linear_term {
    - : 2, unary;
    * : 1, binary, left;
    + : 0, binary, left;
    - : 0, binary, left
    };

    dom_term {
    - : 3, unary;
    + : 3, unary;
    * : 2, binary, left;
    + : 1, binary, left;
    - : 1, binary, left;
    .. : 0, binary, left
    };

    &dom/0 : dom_term, {=}, linear_term, head;
    &sum/0 : linear_term, {<=,>=,>,<,=,!=}, linear_term, any;
    &logic/1 : linear_term, head
}.

\end{lstlisting}

Consider Listing~\ref{list:liatheory}. It introduces the reader to the LIA language specification for grounder \gringo used within \ezsmtv3, which echoes the one utilized within \clingcon version~5 (\clingconv5)\footnote{The theory specification used within \clingconv5 is located at \url{https://github.com/potassco/clingcon/blob/master/libclingcon/clingcon/parsing.hh}\ .} -- the latest version of system \clingcon rooted in the ideas by~\cite{ost17}. Thus, 
any CAS(LIA) program  for \ezsmtv3 can be seen as a program written for \clingconv5 so that it can be solved by that system also.
(It is due to remark that \clingconv5's specification has additions that for instance specify such a directive as the {\tt \&show} statement. Yet, \ezsmtv3 does not support statements of the kind.)
We can see the specification in  Listing~\ref{list:liatheory} as a collection of requirements on the kinds of statements that we expect \gringo to process. We refer the reader to the paper by \cite{gebser2016theory} for more details and intuitions behind the presented theory specification. Here, we utilize examples to illustrate its purpose. In addition to syntactic restrictions on the kinds of statements supported by specifications of Listing~\ref{list:liatheory},
we  pose additional requirements on these expressions, which have to be verified at the level when \gringo output is being processed. Within \ezsmtv3, we adopt the requirements closely related to those  described by \citeauthor{ost17}~(\citeyear[Pages 12 and 13]{ost17}) for the constraints expressed using key words {\tt \&dom} and {\tt \&sum}. 
We now summarize the requirements and
also discuss the nature of these constraints and how they are captured by \ezsmtv3.

\noindent
\textbf{Domain Constraints} have the form
\begin{equation}
\texttt{$\&dom\{d_1; \dots; d_m\} = t$},
\label{eq:domst}
\end{equation}
where: 
\begin{itemize}
    \item $d_i$ ($1\leq i\leq m$) can be  $u$ or a range $v..w$, with $u$, $v$, $w$ being of the form~\eqref{eq:exp} 
    so that 
    \begin{itemize}
        \item $a_i$ and $x_i$ ($1\leq i\leq n$) are integers (with  typical conventions such as, for instance, if one of the coefficients is in the multiplications of this expression is $1$ it can be omitted) and thus, $u$, $v$, and $w$ can be  evaluated to integers); 
     and 
\item         $v\leq w$.   
    \end{itemize}
        \item $t$ is a constraint variable.
\end{itemize}
If expression~\eqref{eq:domst} is such that every $d_i$ ($1\leq i\leq m$) is either  $u$ or a range $v..w$, with $u$, $v$, $w$ being 
{\em integers} we call this statement {\em normal}.
This expression is intuitively understood as imposing the following requirement on values that constraint variable $t$ can be mapped to. Namely, any value in the following set  $\bigcup_{i=1}^{n} [d_i]$, where
\begin{equation*}
    [d] =
    \begin{cases*}
      $\{u\}$ & if $d$ is $u$ \\
      $\{v,..,w\}$ & if $d$ is $v..w$, where $v\leq w$.
    \end{cases*}
  \end{equation*}

Internally, \ezsmtv3  simplifies $\&dom$ statements by evaluating  possibly complex linear expressions occurring in these statements into corresponding integers resulting in the normal $\&dom$ statement. For instance, consider the following lines to occur in some \ezsmtv3 program
\begin{verbatim}
&dom{1..3; 5+3*4} = x:- a, not b.
&dom{1+2..4*4} = x.
\end{verbatim}
Internally, they will be simplified by the system  into
\begin{equation}
\begin{array}{l}
\&dom\{1..3; 17\}=x\ar a,\ not\ b.\\
\&dom\{3..16\} = x.   
\end{array}
\label{eq:exdomst}
\end{equation}

Syntactically, a (ground) \ezsmtv3 rule  containing normal $\&dom$ constraint  has the form
\begin{equation}
D\ar B,
\label{eq:domrule}
\end{equation}
where $D$ is expression~\eqref{eq:domst} and $B$ is the body of this rule understood as  in~\eqref{e:ruleabr}.
Given the fact that within \ezsmtv3 we utilize SMT(LIA) formulas behind the stage to reason over a CAS(LIA) program, we present the semantics of statement~\eqref{eq:domrule} by means of translating it into an SMT(LIA) formula that has to be satisfied whenever statement~\eqref{eq:domrule} appears in the considered program. We view~\eqref{eq:domrule}  as an abbreviation for the following SMT(LIA) implication
$$
B\rightarrow \big([[d_1]]\vee \cdots \vee [[d_n]]\big ),  
$$
where
\begin{equation*}
    [[d]] =
    \begin{cases*}
      \hbox{$t= u$} & if $d$ is $u$ \\
      \hbox{$\big(t\geq v \wedge  t\leq w\big)$} & if $d$ is $v..w$.
    \end{cases*}
  \end{equation*}
Recall that we identify body $B$ with the respective conjunction. When $B$ is empty (as, for instance, in the second line of~\eqref{eq:exdomst}), we can simplify the implication above and identify it with an expression
$$
[[d_1]]\vee \cdots \vee [[d_n]].  
$$

For instance, the ground \ezsmtv3 rules listed in~\eqref{eq:exdomst} are understood as the conjunction of the following SMT(LIA) formulas:
$$
\begin{array}{l}
(a\wedge \neg b)\rar \big((x\geq 1\wedge x\leq 3)\vee x=17\big),\\
(x\geq 3\wedge x\leq 16).
\end{array}
$$

Here, it is due to note that \clingcon is based on finite domain constraint solving so that in its implementation constraint variables over integers are considered within a default 
 domain  $-2^{30}\ ..\  2^{30}$ unless a $\&dom$ expression is provided for this variable that restricts its range; in case of \ezsmtv3 no restrictions on the range of integers are considered by default.



\noindent
\textbf{Linear Constraints} have the form
\begin{equation}
\&sum\{t_1; \dots; t_n\} \bowtie t_{n+1},    \label{eq:sumil}
\end{equation}
where: 
\begin{itemize}
    \item each $t_i$ $(1\leq i\leq n)$ is an integer linear expression\footnote{Within the implementation,   integer linear expressions are understood more liberally than defined here so that, for example $2\times2$ or $(5+2)\times z$ are considered within the realm of allowed syntax.};
    \item 
   $\bowtie$ belongs to \eqref{eq:arithmop}.
\end{itemize}
This syntax captures  expressions of the form $t_1 + t_2 + \dots + t_m \bowtie t_{m+1}$. In turn,  using standard algebraic operations this expression can be transformed into an integer linear constraint. We view~\eqref{eq:sumil} as an irregular atom corresponding to an underlying integer linear constraint (note that there may be multiple equivalent representations of such a constraint and any of these suffice for our purposes; indeed $x<1$ can be seen as an equivalent representation to $x-1\leq 0$).  

For instance, the expression of the form
\begin{verbatim}
&sum{2*2;3+x+(5+2)*z}=y
\end{verbatim}
occurring within a ground \ezsmtv3 program is identified with an irregular atom
$$
|x-y+7\times z=-7|
$$
which has a natural mapping into respective LIA constraint (recall our convention to use vertical bars to denote irregular atoms).

Syntactically, a (ground) \ezsmtv3 rule
may contain expressions of the form~\eqref{eq:sumil} both in the head and the body of the rule, while \ezsmtv3 identifies them with respective irregular atoms.
For instance, the rule of the form
\begin{equation}
\hbox{{\tt \&sum\{2*2;3+x+(5+2)*z\}=y:- a, not b.}}
\label{eq:denial1.1}
\end{equation}
is understood as a denial
\begin{equation}
    \ar \neg |x-y+7z=-7|,\ a,\ not\  b. 
\label{eq:denial1}
\end{equation}

\begin{example}
Recall a program from Example~\ref{example:cas}.
Using the described  \ezsmtv3 CAS(LIA) language this program has the following form
\begin{verbatim}
  {a}.
  b:-a.
  c:-not a.
  :-a, &sum{x}>=12.    
\end{verbatim}
\end{example}

\subsubsection{Traveling Salesman Problem as CAS(LIA) Program}\label{sec:tsp}
Let us state a {\em variant} of the {\em Traveling Salesman Problem}:
\begin{displayquote}
{\em We are given a graph with  nodes as cities and edges as roads. Each road directly connects a pair of cities, and costs a salesman some time to go through (time is expressed as a positive integer value in this variant of the problem). The salesman is supposed to pass each city exactly once.
		 		Find:  {\em a route} traversing  {\em all the cities, yet only once visiting each one of them, under} certain  {\em maximum cost} of total time. }
		 		\end{displayquote}
In the classical formulation of the TS problem, a route with the minimum cost  is of interest. 
Here, we state a decision problem in place of a related optimization problem. Also, in the classical formulation there is no restriction on weights over routes being integers.

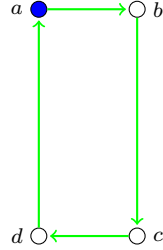
\begin{figure}
\footnotesize	
	\begin{tabular}{l|ll}
		
		An instance  
		with max cost 4&Solution 1&Solution 2\\
		\begin{minipage}[t]{.33\textwidth}
			\flushleft
			\begin{tikzpicture}[shorten >=1pt,node distance=3cm,on grid,auto]
			\tikzstyle{state}=[shape=circle,thick,draw,minimum size=1.5cm]
			\vertexblue (a1) at (2.7,3) [label=left:$a$] {};  
			\vertex (b1) at (4,3) [label=right:$b$] {};
			\vertex (d1) at (2.7,0) [label=left:$d$] {};
			\vertex (c1) at (4,0) [label=right:$c$] {};
			\path[-,draw,thick]
			(a1) edge node {1}  (b1)
			(b1) edge node {1}  (c1)
			(c1) edge node {1}  (d1)
			(d1) edge node {1}  (a1)
			(a1) edge node {2}  (c1)
			(b1) edge node {2}  (d1);
			
			
			\end{tikzpicture}
			
		\end{minipage}
		
		&
		\begin{minipage}[t]{.33\textwidth}
			
			\begin{tikzpicture}[shorten >=1pt,node distance=3cm,on grid,auto]
			\tikzstyle{state}=[shape=circle,thick,draw,minimum size=1.5cm]
			
			\vertexblue (a1) at (2.7,3.5) [label=left:$a$] {};  
			\vertex (b1) at (4,3.5) [label=right:$b$] {};
			\vertex (d1) at (2.7,0.5) [label=left:$d$] {};
			\vertex (c1) at (4,0.5) [label=right:$c$] {};
			\path[->,draw,green,thick]
			(a1) edge   (d1)
			(d1) edge   (c1)
			(c1) edge   (b1)
			(b1) edge   (a1);
			
			
			\end{tikzpicture}
			
		\end{minipage}
  &
		\begin{minipage}[t]{.33\textwidth}
			\flushleft
			\begin{tikzpicture}[shorten >=1pt,node distance=3cm,on grid,auto]
			\tikzstyle{state}=[shape=circle,thick,draw,minimum size=1.5cm]
			
			\vertexblue (a1) at (2.7,3) [label=left:$a$] {};  
			\vertex (b1) at (4,3) [label=right:$b$] {};
			\vertex (d1) at (2.7,0) [label=left:$d$] {};
			\vertex (c1) at (4,0) [label=right:$c$] {};
			\path[->,draw,green,thick]
			(a1) edge   (b1)
			(b1) edge   (c1)
			(c1) edge   (d1)
			(d1) edge   (a1);
			
			
			\end{tikzpicture}
		\end{minipage}
	\end{tabular}
	\normalsize
	\caption{Sample TS Instance and Solutions\label{fig:ts}}		 		
\end{figure}

Figure~\ref{fig:ts} shows an instance of the TS problem (a weighted graph).
Listing~\ref{list:tsm-inst} encodes this representation as a set of facts. 
On the right hand side  of Figure~\ref{fig:ts}, we find two solutions to this problem.

\begin{lstlisting}[
  caption = {Encoding of the TS Instance. 
%   The program is annotated with \emph{achievements}.
},
  label={list:tsm-inst},
  basicstyle=\ttfamily\small,
%  numbers=left,
%  stepnumber=2,
]
city(a). city(b). city(c). city(d).
initial(a).
road(a,b).   road(b,c).   road(c,d).
cost(a,b,1). cost(b,c,1). cost(c,d,1). 
road(d,a).   road(a,c).   road(b,d).  
cost(d,a,1). cost(a,c,2). cost(b,d,2).  
maxCost(4).
\end{lstlisting}


Listing~\ref{list:tsm} presents the CASP encoding  for the TS problem, whose instances are provided in the style of the instance presented in Listing~\ref{list:tsm-inst}. This CASP encoding respects the LIA logic and supports the syntax specified by the theory specification in Listing~\ref{list:liatheory}. 
Let us start by stating intuitions behind this encoding.
The first line specifies that the {\tt road} relation is symmetric. The second line suggests that the {\tt cost} of the road taken in either directions is the same. Line 4 specifies that for each city in the problem exactly one road that leads away from the city has to be part of the solution encoded by binary relation {\tt route}.
Line 5 specifies that for each city exactly one road that leads into this city has to be part of the solution. Lines 7 and 8 encode a notion of a reached city from an initial city. Line 10 requires that each city in the problem is identified as reached from the initial city. 
Lines~12 through~15 utilize constructs, whose syntax is defined within the theory specification in Listing~\ref{list:liatheory}. In other words, in the absence of the code within Listing~\ref{list:liatheory},  \gringo would identify these lines as outside of the scope of its applicability. Line~12 
\begin{itemize}
\item introduces irregular atoms into discourse -- atoms that refer to constraint variables and
\item specifies  possible  values for these constraint variables.
\end{itemize}
Lines 13 through~15 state  requirements/constraints on these variables.
In particular, Line 12 specifies a domain of possible values for the instances of  constraint variables of the form {\tt c(X,Y)}. Namely, their domains are restricted  by two values: one being $0$ and another being the costs associated with the roads from $X$ to $Y$. Lines 13 and~14 state the conditions on when instances of  constraint variables {\tt c(X,Y)} are assigned $0$ or the associated cost. Line 15 specifies an integer linear constraint that states that the sum of all possible instances of constraint variables {\tt c(X,Y)} should not exceed the maximum cost specified.

\begin{lstlisting}[
  caption = {Encoding of the TS Problem. 
%   The program is annotated with \emph{achievements}.
},
  label={list:tsm},
 % float=tp,
  basicstyle=\ttfamily\small,
  numbers=left,
  stepnumber=2,
]
road(Y,X):-road(X,Y).
cost(Y,X,C):-cost(X,Y,C).

1{route(X,Y): road(X,Y)}1:-city(X).
1{route(X,Y): road(X,Y)}1:-city(Y).

reached(X):-initial(X).
reached(Y):-reached(X), route(X,Y).

:-city(X), not reached(X).

&dom {0;C} = c(X,Y) :- cost(X,Y,C).
&sum {c(X,Y)} =0 :-cost(X,Y,C), not route(X,Y).
&sum {c(X,Y)} =C :-cost(X,Y,C), route(X,Y).
:- &sum {c(X,Y):cost(X,Y,C)} > W, maxCost(W).

\end{lstlisting}

It is easy to see that the considered encoding of the  TS problem  contains  kinds of rules that are outside of the  syntax of logic programs presented in the Background section. In particular, this program uses
\begin{itemize}
\item ASP variables --- namely, $X$, $Y$, $C$, and $W$ (identifiers starting with the capital letters) --- so that program's atoms are not propositional;
\item  aggregate expressions within Lines 4 and 5. In fact, each of these rules is an abbreviation for two rules, where one rule contains a choice expression in the head and another rule is a constraint containing count-aggregate expression in the body.
\end{itemize}

Aggregate expressions are the common constructs within the practice of answer set programming. We refer an interested reader to the work by~\cite{CalimeriFGIKKLRS}, for instance, for more formal details on aggregates. Here, we informally discuss their roles using  Line 4 within Listing~\ref{list:tsm} as an example. The expression presented on the line is an abbreviation for two rules:
\begin{verbatim}
{route(X,Y)}:- road(X,Y), city(X).
:- not #count{X,Y:route(X,Y),road(X,Y)}=1, city(X).
\end{verbatim}
The first line can be intuitively read as {\em any road leading from some city {\em may form} a part of the route}.
The word {\em may} points at the {\em choice}. 
The second rule contains a {\em count}-aggregate expression and states that {\em for a city exactly one tuple corresponding to a road should be considered to be part of the route}.

This is a good place to demystify the effects of the grounding process and the role of ASP variables. 
The process of {\em grounding} is defined through 
ensuring  that ASP variables are instantiated with all possible permutations of the object constants, so that a rule with ASP variables can be seen as an abbreviation for the group of propositional rules instantiated with the object constants occurring in the program.  
Grounder \gringo performs a process denoted as {\em intelligent grounding} that is similar to a procedure well described by~\cite{fab12}. While performing intelligent grounding a system attempts not only to instantiate given logic rules with all possible object constants of the considered program, but also to perform some simplifications and reductions that still guarantee that the produced propositional program has the same answer sets as the one that would be produced by the straight-forward instantiation of grounding.
 The exact procedure behind {\gringo} is best documented by~\cite{kaminskiPhD}.
Lines in Listings~\ref{list:tsm-inst} and~\ref{list:grd}  form the output of \gringo,
when it is invoked with the flag {\em -t} on the code obtained by concatenating the lines within Listings~\ref{list:liatheory},~\ref{list:tsm-inst} and~\ref{list:tsm}. 
Flag {\em -t} instructs \gringo to print output in human readable form.
Let us now  discuss intuitions on which snippets of code within Listings~\ref{list:tsm-inst} and \ref{list:tsm} are relevant in producing propositional rules in Listing~\ref{list:grd}:
\begin{itemize}[leftmargin=3.5mm]
\item Lines 1 through 4 are produced by \gringo by relying on the facts in Listing~\ref{list:tsm-inst} and Lines 1 and 2 in Listing~\ref{list:tsm}. 
\item Lines 5-20 are produced by \gringo by relying on the facts in Listing~\ref{list:tsm-inst}; Lines 1 and 3 in Listing~\ref{list:grd}; and Lines 4 and 5 in Listing~\ref{list:tsm}. 
\item Lines 21-28 are produced by \gringo by relying on the fact in Line 2 in Listing~\ref{list:tsm-inst}; Lines 5-20 in Listing~\ref{list:grd} suggesting which tuples may appear in {\tt route} relation; and Lines 7 and 8 in Listing~\ref{list:tsm}. 
\item Lines 30-35 are due to Line 12  in Listing~\ref{list:tsm} and the {\tt cost} relations established in Listing~\ref{list:tsm-inst} and Lines 2 and 4 in Listing~\ref{list:grd}.
\item Lines 30-35 are due to Line 12  in Listing~\ref{list:tsm} and the {\tt cost} relations established in Listing~\ref{list:tsm-inst} and Lines 2 and 4 in Listing~\ref{list:grd}.
\item Lines 36-53 are due to Lines 13 and 14  in Listing~\ref{list:tsm}; the {\tt cost} relations established in Listing~\ref{list:tsm-inst} and Lines 2 and 4 in Listing~\ref{list:grd}; and Lines 5-20 in Listing~\ref{list:grd}. 
\item Line 54 is due to Line 15   in Listing~\ref{list:tsm}; the {\tt maxCost} given in Listing~\ref{list:tsm-inst};
the {\tt cost} relations established in Listing~\ref{list:tsm-inst} and Lines 2,  4 in Listing~\ref{list:grd}.
\end{itemize}

\begin{lstlisting}[
  caption = {Part of the Grounded TS Problem with respect to TS Instance in Listing~\ref{list:tsm-inst} 
%   The program is annotated with \emph{achievements}.
},
float=tp,
numberfirstline = false,
  label={list:grd},
  basicstyle=\ttfamily\footnotesize,
  numbers=left,
  stepnumber=2,
]
road(d,b).    road(c,a).    road(a,d).
cost(d,b,2).  cost(c,a,2).  cost(a,d,1).
road(d,c).    road(c,b).    road(b,a).
cost(d,c,1).  cost(c,b,1).  cost(b,a,1).
1<=#count{0,route(a,b):route(a,b);0,route(a,c):route(a,c);
                                  0,route(a,d):route(a,d)}<=1.
1<=#count{0,route(b,c):route(b,c);0,route(b,d):route(b,d);
                                  0,route(b,a):route(b,a)}<=1.
1<=#count{0,route(c,d):route(c,d);0,route(c,a):route(c,a);
                                  0,route(c,b):route(c,b)}<=1.
1<=#count{0,route(d,a):route(d,a);0,route(d,b):route(d,b);
                                  0,route(d,c):route(d,c)}<=1.
1<=#count{0,route(d,a):route(d,a);0,route(c,a):route(c,a);
                                  0,route(b,a):route(b,a)}<=1.
1<=#count{0,route(a,b):route(a,b);0,route(d,b):route(d,b);
                                  0,route(c,b):route(c,b)}<=1.
1<=#count{0,route(b,c):route(b,c);0,route(a,c):route(a,c);
                                  0,route(d,c):route(d,c)}<=1.
1<=#count{0,route(c,d):route(c,d);0,route(b,d):route(b,d);
                                  0,route(a,d):route(a,d)}<=1.
reached(a).              reached(b):-route(a,b).
reached(c):-route(a,c).  reached(d):-route(a,d).
reached(b):-route(d,b),reached(d).  
reached(c):-route(d,c),reached(d).
reached(d):-route(c,d),reached(c).
reached(b):-route(c,b),reached(c).
reached(c):-route(b,c),reached(b).
reached(d):-route(b,d),reached(b).
:-not reached(b).   :-not reached(c).   :-not reached(d).
&dom{(0;1)}=(c(a,b)).   &dom{(0;1)}=(c(b,c)). 
&dom{(0;1)}=(c(c,d)).   &dom{(0;1)}=(c(d,a)).
&dom{(0;2)}=(c(a,c)).   &dom{(0;2)}=(c(b,d)).
&dom{(0;2)}=(c(d,b)).   &dom{(0;2)}=(c(c,a)).
&dom{(0;1)}=(c(a,d)).   &dom{(0;1)}=(c(d,c)).
&dom{(0;1)}=(c(c,b)).   &dom{(0;1)}=(c(b,a)).
&sum{c(a,b)}=(0):-not route(a,b).  
&sum{c(b,c)}=(0):-not route(b,c).
&sum{c(c,d)}=(0):-not route(c,d).
&sum{c(d,a)}=(0):-not route(d,a).
&sum{c(a,c)}=(0):-not route(a,c).
&sum{c(b,d)}=(0):-not route(b,d).
&sum{c(d,b)}=(0):-not route(d,b).
&sum{c(c,a)}=(0):-not route(c,a).
&sum{c(a,d)}=(0):-not route(a,d).
&sum{c(d,c)}=(0):-not route(d,c).
&sum{c(c,b)}=(0):-not route(c,b).  
&sum{c(b,a)}=(0):-not route(b,a).
&sum{c(a,b)}=(1):-route(a,b).  &sum{c(b,c)}=(1):-route(b,c).
&sum{c(c,d)}=(1):-route(c,d).  &sum{c(d,a)}=(1):-route(d,a). 
&sum{c(a,c)}=(2):-route(a,c).  &sum{c(b,d)}=(2):-route(b,d).
&sum{c(d,b)}=(2):-route(d,b).  &sum{c(c,a)}=(2):-route(c,a).
&sum{c(a,d)}=(1):-route(a,d).  &sum{c(d,c)}=(1):-route(d,c).
&sum{c(c,b)}=(1):-route(c,b).  &sum{c(b,a)}=(1):-route(b,a).
:-&sum{c(a,b); c(b,c); c(c,d); c(d,a); c(a,c); c(b,d); 
       c(d,b); c(c,a); c(a,d); c(d,c); c(c,b); c(b,a)}>(4).
\end{lstlisting}

\paragraph{How rules with {\tt \&sum} and ASP variables connect to CAS(LIA) rules}
Listing~\ref{list:grd} provides us with the propositional rendering of CAS program encoding of our running example of the TS problem. Let us use it to connect to the formal notions introduced in Section~\ref{sec:background}.
Consider rules in Lines 36 and 54-55 in  Listing~\ref{list:grd}. We can view these 
as corresponding to the following two rules
written in the syntax discussed in Section~\ref{sec:background} 
$$
\begin{array}{rl}
     &\mid c(a,b)=0\mid \ar not\ route(a,b). \\
     &\ar \mid c(a,b) +c(b,c)+ c(c,d)+ c(d,a)+ c(a,c)+ c(b,d)+ \\
     &~~~~~c(d,b)+ c(c,a)+ c(a,d)+ c(d,c)+ c(c,b)+ c(b,a)>4\mid.
\end{array}
$$
In these rules,
two irregular atoms appear marked by vertical bars. They naturally translate into LIA constraints with twelve integer constraint variables including 
$c(a,b)$ and $c(b,c)$, for example.


\paragraph{Invoking \ezsmtv3}
A unique capability of \ezsmtv3 lies in the fact that it provides a frontend to distinct SMT solvers, namely, \cvcFour, \cvcFive, \yices, and \zThree. 
One may specify an SMT solver of interest at the command line. In addition, one may specify whether 
single or multiple 
 answer sets (or  extended answer sets)  are of interest. 

Let us assume the presence of the files
\begin{enumerate}
    \item {\tt tsp.inst} -- whose content is present in  Listing~\ref{list:tsm-inst};
\item {\tt tsp.enc} -- whose content is present in  Listing~\ref{list:tsm} together with an additional directive of the form {\tt \#show route/2.} This directive instructs the system to only display atoms formed with this predicate symbol in the output.
\end{enumerate}
Then, the command line

{\tt ezsmt  tsp.inst tsp.enc -s z3 -e 0 -E}

\noindent 
produces the output given in Listing~\ref{list:tsm-output}.
This output matches the solutions listed in Figure~\ref{fig:ts}: Answer 1 and 2 encode Solutions 1 and 2, respectively.
Within this command line in addition to specifying files containing the program to process, we state 
\begin{itemize}
\item a backend SMT solver that should be used -- here, {\zThree} -- with {\tt  -s z3},
\item a number of answer sets that should be enumerated -- here, {\em all} --  with {\tt  -e 0},
\item a request to consider extended answer sets within the enumeration process {\tt  -E}.
\end{itemize}
\begin{lstlisting}[
  caption = {\ezsmtv3 output for the TS sample problem. 
%   The program is annotated with \emph{achievements}.
},
float=tp,
numberfirstline = false,
  label={list:tsm-output},
  basicstyle=\ttfamily\small,
%  numbers=left,
%  stepnumber=2,
]
Answer 1: route(a,d) route(d,c) route(c,b) route(b,a)
c(a,b)=0 c(b,c)=0 c(c,d)=0 c(d,a)=0 c(a,c)=0 c(b,d)=0 
c(d,b)=0 c(c,a)=0 c(a,d)=1 c(d,c)=1 c(c,b)=1 c(b,a)=1
Finished round 1 in 118ms
  0ms SMT Check Satisfiability
  16ms SMT Get Values

Answer 2: route(a,b) route(b,c) route(c,d) route(d,a)
c(a,b)=1 c(b,c)=1 c(c,d)=1 c(d,a)=1 c(a,c)=0 c(b,d)=0 
c(d,b)=0 c(c,a)=0 c(a,d)=0 c(d,c)=0 c(c,b)=0 c(b,a)=0
Finished round 2 in 67ms
  0ms SMT Check Satisfiability
  35ms SMT Get Values
\end{lstlisting}

It is due to remark that within the code base of \ezsmtv3, the specifications presented in Listing~\ref{list:liatheory} are used to instruct \gringo (invoked within) on what expressions it should find syntactically valid (Section~\ref{sec:arch} narrates the details on the architecture of \ezsmtv3). 

Let us now speak about the distinction between  {\tt  -e 0} and {\tt -e 0  -E} settings. The former is concerned with enumerating distinct answer sets disregarding the specific values that constraint variables obtain. The later will instruct \ezsmtv3 to enumerate distinct extended answer sets. Let us consider a simple program presented in Listing~\ref{list:many}.
The  \ezsmtv3 system invoked with \hbox{{\tt -e 0}} on this sample program produces two solutions total that correspond to distinct answer sets, 
while  \ezsmtv3 invoked with {\tt -e 0 -E}  produces three extended answer sets. 
\begin{lstlisting}[
  caption = {Sample CAS(LIA) \ezsmtv3 program with multiple (extended) answer sets
%   The program is annotated with \emph{achievements}.
},
float=tp,
numberfirstline = false,
  label={list:many},
  basicstyle=\ttfamily\small,
%  numbers=left,
%  stepnumber=2,
]
&dom{1..3}=x.
{a}.
&sum{x}=1:- a.
&sum{x}<3:- not a.
\end{lstlisting}

\subsubsection{\ezsmtv3 CAS(LRA),  CAS(LIRA), and CAS(IDL) Languages}\label{sec:ezsmtl-s}
We now present the details on the encodings of the constraints supported by the \ezsmtv3 system when it assumes the roles of CAS(LRA),  CAS(LIRA), and CAS(IDL) solvers, respectively.

\paragraph{The CAS(LRA) Language}
Section~\ref{sec:caslialang} described the CAS(LIA) language supported by \ezsmtv3. The same section can be seen as the one describing the details of the CAS(LRA) language supported by the system modulo the condition that non-integer real numbers are listed using quotation marks.  
For instance, the expression of the form
\begin{equation}
\hbox{
{\tt \&sum\{"2.4"*2;3+x+(5+2)*z\}=y}}
\label{ex:lra}
\end{equation}
occurring within a ground CAS(LRA) \ezsmtv3 program is identified with an irregular atom 
\begin{equation}
|x-y+7\times z=-7.8|
\label{ex:lra2}
\end{equation}
which has a natural mapping into a respective LRA constraint with three constraint variables over reals, namely, $x$, $y$, and $z$.
Just as we attempted to make the fragment of the   CAS(LIA) language supported by \ezsmtv3 compatible with the \clingcon language, we also 
attempted to make the fragment of the   CAS(LRA) language supported by \ezsmtv3 compatible with the \clingolp~\cite[]{jan17} language so that a CAS(LRA) program written for \ezsmtv3 can be processed by \clingolp system (modulo omitting the directive $\&logic(lra).$ described below). It is due to note that
\begin{itemize}
\item \clingolp supports an additional optimization statement that is outside of the scope of \ezsmtv3  and 
\item \clingolp is less permissive in the form of the $\&sum$ statements it allows. For example, the expression of the form~\eqref{ex:lra} is considered syntactically invalid  by \clingolp. Yet, recall how this expression corresponds to irregular atom~\eqref{ex:lra2}, which we could encode in the syntax understood by \clingolp as follows
$$
\hbox{{\tt \&sum\{x;(-1)* y;7* z\}="-7.8".}}
$$
\end{itemize}
In addition, an \ezsmtv3 CAS(LRA) program may contain the following declaration
\begin{verbatim}
&logic(lra).
\end{verbatim}
This declaration instructs \ezsmtv3 that the program it is dealing with is CAS(LRA) program. Alternatively, a flag {\tt -l 1} within the command line can be used to invoke \ezsmtv3 instructing it to process a CAS(LRA) program. 

The theory specification for CAS(LRA)  
is identical to the specification in Listing~\ref{list:liatheory} modulo an additional line inserted after line 18 of that listing:
\beq
\hbox{\tt \&logic/1 : var\_term, head;}
\eeq{eq:logicstate}
This additional line allows \ezsmtv3 to introduce the directive {\tt  \&logic(lra)}.

\paragraph{The CAS(LIRA) Language}

The theory specification for CAS(LIRA) programs required by \gringo is identical to the specification listed in Listing~\ref{list:liatheory} modulo two additional lines inserted after line 18 of that listing. The first line is presented in~\eqref{eq:logicstate} and the second line follows:
$$
\hbox{\tt \&type/2: var\_term, head;}
$$
These two additional lines
allow \ezsmtv3 to process the directives of the following kind
\begin{equation}
\begin{array}{l}
\hbox{{\tt \&logic(lira).}}\\
\hbox{{\tt \&type\{x; y\}=int.}}\\
\end{array}    
\label{eq:specs_auflira}
\end{equation}
In this snippet of sample code, the first line declares to \ezsmtv3 that the program it currently considers is within the CAS(LIRA) language;
alternatively, a user may use flag {\tt -l 2} within the command line to invoke \ezsmtv3 in such a mode.
Before we discuss the role of the second line, let us introduce a term {\em functional name} of a constraint variable. Within the programs that \ezsmtv3 supports a constraint variable may take one of two forms
$$
\begin{array}{l}
v\\
v(t_1,\dots,t_n).
\end{array}
$$
In these expressions, we call $v$ a {\em functional name} of a constraint variable.
The sample code {\tt \&type\{x; y\}=int.}
states a condition that any constraint variable with functional name $x$ or~$y$  
occurring in a given program is considered to be integer.
Any constraint variable occurring within a program whose functional name is missing from a declaration of this kind is considered to be a constraint variable over reals.

For the remainder, Section~\ref{sec:caslialang} can be seen as a section describing the details of the CAS(LIRA) language supported by \ezsmtv3 modulo the condition that real numbers that are not integers are listed using quotation marks.  
For instance, expression
\eqref{ex:lra}
occurring within a ground CAS(LIRA) \ezsmtv3 program that contains lines in \eqref{eq:specs_auflira} and no other type-declarations is identified with an irregular atom of the form~\eqref{ex:lra2},
which has a natural mapping into the respective LIRA constraint with  integer constraint variables $x$ and $y$,  and real constraint variable $z$.

\paragraph{The CAS(IDL) Language}
The theory specification for CAS(IDL) programs  is  in Listing~\ref{list:idltheory}. 
This specification allows \ezsmtv3 to provide support for
\begin{itemize}
\item  difference logic constraints of the form~\eqref{eq:idlcon}; and
\item the declaration
$$
\hbox{{\tt  \&logic(idl).}}
$$ This directive  instructs \ezsmtv3 that it is dealing with CAS(IDL) program; alternatively, a user may use flag {\tt -l 3} for the same instruction.
\end{itemize}
\begin{lstlisting}[
  caption = {Encoding of IDL Logic in \gringo version 5. 
%   The program is annotated with \emph{achievements}.
},
  label={list:idltheory},  basicstyle=\ttfamily\small,
%  numbers=left,
%  stepnumber=2,
]
#theory idl {
    linear_term {
    - : 2, unary;
    * : 1, binary, left;
    + : 0, binary, left;
    - : 0, binary, left
    };

    dom_term {
    - : 3, unary;
    + : 3, unary;
    * : 2, binary, left;
    + : 1, binary, left;
    - : 1, binary, left;
    .. : 0, binary, left
    };

    &dom/0 : dom_term, {=}, linear_term, head;
    &diff/0 : linear_term, {<=,>=,<,>,=,!=}, linear_term, any;
    &logic/1 : linear_term, head
}.
\end{lstlisting}

For instance, expressions of the form
\begin{equation}
\hbox{\tt \&diff\{x-y\}<=5}
\label{eq:difat1}
\end{equation}
and
\begin{equation}
\hbox{\tt \&diff\{x\}<y}
\label{eq:difat2}
\end{equation}
occurring within a ground CAS(LRA) \ezsmtv3 program are identified with  irregular atoms 
$$
|x-y\leq 5|
$$
and
$$
|x<y|,
$$
respectively. Both of these irregular atoms have  a natural mapping into respective IDL constraints.

It is due to note that the CAS(IDL) \ezsmtv3 program is often suitable for processing with solver \clingodl~\citep{jan17}. Yet, the dialect of  CAS(IDL) \ezsmtv3 programs permits the following expressions that are outside the language fragment of \clingodl:
\begin{itemize}
\item the $\&logic$ directive, which specifies the IDL logic to be used within the encoding. This directive can be eliminated from programs when proper flag is used to invoke \ezsmtv3.
\item the $\&dom$ specifications for variables. These expressions are treated in the same way as described for the case of CAS(LIA) fragment, and it is easy to see that the resulting SMT formulas are within the  SMT(IDL) fragment. \clingodl bypasses the support for this language feature.     
\end{itemize}

\subsection{\ezsmtv3 Architecture}\label{sec:arch}

\begin{figure}[H]
    \centering
    \includegraphics[width=0.9\linewidth]{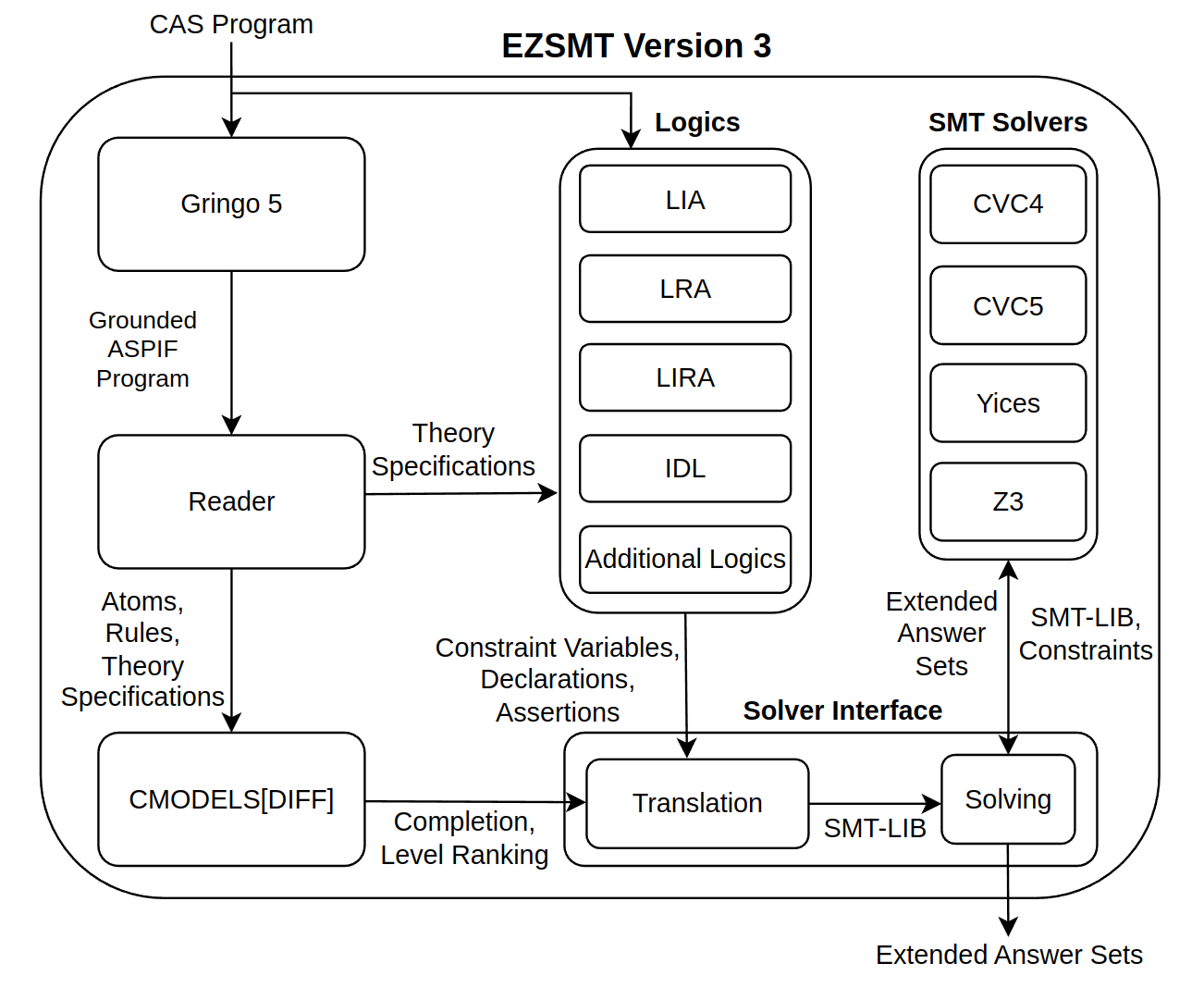}
    \caption{\ezsmtv3 Architecture}
    \label{fig:fig1}
\end{figure}

Figure \ref{fig:fig1} presents the architecture of the \ezsmtv3 system. This system is able to process CAS(LIA), CAS(LRA), CAS(LIRA), and CAS(IDL)
utilizing the language constructs as specified in Section~\ref{sec:languc}. 
We start by briefly describing the system's workings. Then we provide details for its more complex elements. 

At first, the \ezsmtv3 system determines which kind of program it is given -- CAS(LIA), CAS(LRA), CAS(LIRA), or CAS(IDL). After that, it utilizes grounder \gringo \citeayy{gebser2016theory}{kaminski2023build} to eliminate ASP variables. System \gringo produces a ground/propositional program in the format called
Answer Set Programming Intermediate Format (ASPIF)~\citeayy{gebser2016theory}{kaminski2023build}.
The grounded program written in ASPIF  is then read by the Reader component of the system, which stores the rules, regular and irregular  atoms from the program accordingly. The logic interface is then set and the corresponding constraint variables are declared with specified types. Routines of system \cmodelsdiff~\citep{shen18} are used to compute completion and level rankings of the program. Then, the \ezsmtv3 system translates the completion augmented with level rankings into SMT formulas in the syntax of the standard SMT-LIB language~\citep{BarST-SMTLIB}. These formulas are then fed into an SMT solver, which finds a model of the formulas. Each found model corresponds to an extended answer set of the given program. We now provide more essential details behind each sub-component of the \ezsmtv3 system depicted in Figure \ref{fig:fig1}.

\paragraph{The \gringo 5 block}
Section~\ref{sec:tsp} used an instance of a  Traveling Salesman problem formalized as CAS(LIA) program to illustrate the process of grounding. Within the \ezsmtv3 grounder
\gringo version 5 is utilized. It is a sub-component of already mentioned system \clingov5.
Sections~\ref{sec:caslialang} and~\ref{sec:ezsmtl-s} highlighted the presence and importance of theory specifications -- recall, Listings~\ref{list:liatheory} and~\ref{list:idltheory} -- 
that enable us to define the syntactic constructs of 
various CAS languages that \gringo is then able to process. For example, the theory specification in Listing~\ref{list:idltheory} instructs \gringo that atoms of the form~\eqref{eq:difat1} and/or~\eqref{eq:difat2} are valid constructs syntactically. Given such a theory specification,  \gringo is able to ground the respective programs and encode these using the ASPIF format~\citeayy{gebser2016theory}{kaminski2023build}.
This format is best documented in the appendix of the extended version of the  paper by~\cite{gebser2016theory}   available at \url{https://www.cs.uni-potsdam.de/wv/publications/DBLP_conf/iclp/GebserKKOSW16x.pdf} .
The grounded logic program in ASPIF can be seen as a list of statements in normal form that utilize numerical values to represent program's atoms and its internal structure.
\paragraph{The Reader block}
The ASPIF statements generated by  grounder \gringo are interpreted by the Reader block. 
The Reader block parses ASPIF statements provided by \gringo and stores the information about rules, regular, irregular atoms, constraint variables into the internal data structures of \ezsmtv3. 

It is due to note that \gringo may recognize some expressions that are outside of the considered syntax as valid. For instance, in the realm of theory specification for IDL constraints presented in Listings~\ref{list:idltheory}, \gringo will recognize
the following expression as a valid irregular atom 
\begin{equation*}
\hbox{\tt \&diff\{x-y\}<=z}   
\end{equation*}
if it occurs within head or body of some of its rules.
Intuitively this expression maps into  
$$x-y<z$$ 
that is outside the syntax of IDL constraints. 
For this reason the Reader block  also implements additional checks to warn the user about mistakes in  the considered encodings.

\paragraph{The Logics block}
Within the Logics block of \ezsmtv3, we first determine whether a given program is of kind CAS(LIA), CAS(LRA), CAS(LIRA), or CAS(IDL).
For that purpose $\&logic$ and/or command line $-l$ directives are used as specified in Section~\ref{sec:languc}. In case of conflicting information between directives expressed by $\&logic$ statement within the considered program and command line, the  $\&logic$ statement has higher precedence. By default, in the absence of any $\&logic$ statement or flag $-l$ in the command line the program is considered to be within the CAS(IDL) fragment. 
Using the information about a kind of a given program, \ezsmtv3 is able to assign each constraint variable occurring within irregular atoms a proper domain. 



\paragraph{The \cmodelsdiff block}
The \ezsmtv3 system incorporates a number of routines stemming from the answer set solver called \cmodelsdiff~\citep{shen18}. In particular, it borrows the \cmodelsdiff code that determines whether a given program is tight,
performs so called completion on a given program, 
computes level ranking formulas in case a formula is not tight, and 
clausifies the resulting formulas.
These routines are key to implementing the translation provided in the concluding part of Section~\ref{sec:casp}. Indeed, bullets 1 and 2 of that translation are captured  by the process of completion and construction of level ranking formulas.
Just as in the case of \cmodelsdiff, we can instruct \ezsmtv3 to construct different kinds of level ranking formulas using flags {\tt -levelRanking}, {\tt -levelRankingStrong}, {\tt -SCClevelRanking}, and {\tt -SCClevelRankingStrong}. We refer the reader to the work by~\cite{shen18} for more details on the different kinds of level ranking formulas supported by \cmodelsdiff and \ezsmtv3.
The default behavior of system \ezsmtv3 is captured by {\tt -SCClevelRanking}.

By default, we set the upper bound for a level ranking variable corresponding to an atom as the number of atoms inside the strongly connected component of the program's dependency graph containing the corresponding atom. A larger upper bound can also be selected using the flag {\tt--all-atoms-upper-bound} which sets the upper bound as the total number of atoms inside the program.
Finally, the resulting formulas are stored  in semi-Dimacs format documented by \cite{susman2016smt}.

\paragraph{The Solver Interface block}
The Solver Interface block  is responsible for two tasks, namely,  translation and  solving. 
In the translation phase, the formulas in semi-Dimacs form obtained from a previous block   are transformed into the syntax of the Standard language for SMT solvers called SMT-LIB ~\citep{BarST-SMTLIB}. The translation procedure is in the style of the one described by \cite{susman2016smt}. During this transformation, in addition to encoding the SMT formula corresponding to a given program and computed in the \cmodelsdiff block, the declarations for the used
SMT logic, propositional atoms, and 
constraint variables are included.

Let us consider a simple example to illustrate transformations occurring within \ezsmtv3.
Assume some CAS(LIA) program that contains
rule~\eqref{eq:denial1.1}, which we understand as a denial~\eqref{eq:denial1}. The \cmodelsdiff block will turn this denial into a group of SMT(LIA) formulas, namely, 
$$
\begin{array}{l}
|x-y+7z=-7|\vee \neg a \vee  b,\\
|x-y+7z=-7|\leftrightarrow x-y+7z=-7     .
\end{array}
$$
Within an SMT-LIB code for the considered program we will find the following lines that we annotate with the comments for readability (comments start with semicolon):
\small
\begin{verbatim}
; Quantifier free Linear Arithmetic: SMT(LIA) language 
(set-logic QF_LIA)

; Declaration of boolean and integer variables used
(declare-fun a () Bool)
(declare-fun b () Bool)
(declare-fun x () Int)
(declare-fun y () Int)
(declare-fun z () Int)

; |EZSMT_THEORY(4)| is the name given within Ezsmtv3 
; to irregular atom |x-y+7z=-7|
(declare-fun |EZSMT_THEORY(4)| () Bool)

; SMT(LIA) formulas stated above encoded in SMT-LIB
(assert (or |EZSMT_THEORY(4)| (not a) b))
(assert (= |EZSMT_THEORY(4)| (= (+ x (* (- 1) y) (* 7 z)) (- 7))))
\end{verbatim}
\normalsize
Figure~\ref{fig:castosmt} summarized what kind of CAS programs are mapped into what kind of SMT formulas.
The table below details a logic declaration statement in SMT-LIB format appropriate to invoke a correct solving routine for the respective SMT formula:

\smallskip
\begin{tabular}{l|l}
    SMT(LIA)& {\tt (set-logic QF\_LIA)} \\
    SMT(LRA)& {\tt (set-logic QF\_LRA)} \\
    SMT(IDL)& {\tt (set-logic QF\_IDL)} \\
    SMT(LIRA)~~~& {\tt (set-logic AUFLIRA)} 
    \end{tabular}
\smallskip



In the solving phase,  an SMT solver is given an obtained SMT-LIB theory. The back-and-forth communication between the Solver Interface block and the SMT solver of choice makes it possible for the system to output multiple (extended) answer sets. The command line directives can be used to provide \ezsmtv3 with a specific number of solutions to be computed.

All SMT solvers currently supported by  the \ezsmtv3 system, namely \cvcFour,   \cvcFive,  \yices, and \zThree implement so called incremental solving. Within incremental solving settings one may invoke an SMT solver on a theory and instruct it to compute a model; once that model is computed the SMT solver puts its computation on hold and waits for further instructions. At that point it is possible to add more assertions to the theory already populated within the  SMT solver's data structures and ask it to look for yet another model of an updated theory. This process can be repeated. 
\ezsmtv3 utilizes incremental solving for computing multiple answer sets (or extended answer sets) by iteratively adding an assertion that is false when previously computed (extended) answer set holds. Given an answer set $A$ of CAS program $P$ the negation of the following formula
\beq
\bigwedge_{atom~a\in A} a~\wedge~  \bigwedge_{a~occurs~in~P~and~a \not\in A} \neg a
\eeq{eq:conjas}
forms such an assertion.
Given an extended answer set $\langle A,\nu\rangle$ of CAS program~$P$, the negation of the formula formed by the conjunction of formulas~\eqref{eq:conjas} and 
$$
\bigwedge_{constraint~variable~x~occurs~in~P} x=\nu(x)
$$
forms such an assertion.

This way of implementing enumeration of multiple (extended) answer sets is inspired by the enumeration done by answer set solver \cmodelsdiff \cite[Section~5]{shen18}. Yet, utilization of incremental mode of SMT solving is unique to \ezsmtv3. The \ezsmtPlus system invoked SMT solvers from scratch each iteration. 


\paragraph{The SMT Solvers block}
In our work, we implemented support within \ezsmtv3 for four SMT solvers, namely, \cvcFour  \citep{barrett2011cvc4}, \cvcFive \citep{barrettcvc5},  \yices \citep{dutertre2006yices}, and \zThree \citep{de2008z3}. 
Yet, given that we use SMT-LIB to interface these solvers it requires limited effort to implement support for any other solver supporting SMT-LIB format. This implementation effort mainly has to be directed towards processing output of the solvers as their output formats are not identical.

\section{Optimizations}\label{sec:optimizations}
We now turn our attention to weak constraints/optimizations supported by \ezsmtv3. It is due to note that such language constructs were out of scope for the system's predecessor \ezsmtPlus.
The \ezsmtv3 system supports the syntax of weak constraints as they are described by~\cite{cal20} as part of the ASP-Core2 standard language of logic programs. Here we provide the natural extension of the semantics of these statements to the CAS programs.

\cite{cal20}   present the syntax of weak constraints allowing ASP variables in the context; then,  grounding is used to obtain propositional program with weak constraints and the notion of an optimal answer set is defined.
Here, we present all relevant definitions using the propositional case but note that \ezsmtv3 provides support for non-grounded statements that are tackled by the means of grounder \gringo.

        A {\em weak constraint} has the form
\beq 
\wr a_1,\dotsc, a_j,\ not\  a_{j+1},\dotsc,\ not\  a_m[w@\ell,t_1,\dots,t_n],
\eeq{eq:wc}
where $m>0$ and $a_1,\ldots,a_m$ are atoms,  $w$ (weight) is an integer, $\ell$ (level) is a positive integer, $t_i$ ($n\geq 0$) are symbols. 
In the sequel, we abbreviate expression
\beq
\wr  a_1,\dotsc, a_j,\ not\  a_{j+1},\dotsc,\ not\  a_m
\eeq{eq:wcbody}
occurring in~\eqref{eq:wc} as $D$ and identify it with the propositional formula
\beq a_1\wedge\dotsc\wedge a_j\wedge\ \neg  a_{j+1}\wedge\dotsc\wedge\ \neg  a_m.
\eeq{eq:wcbodyf}
We may refer to this formula as the {\em body} of a weak constraint.
\begin{definition}[Optimization program or o-program]
An {\em optimization program} (or {\em o-program})  over vocabulary $\sigma$ is a pair~$(P,W)$, where $P$ is a CAS program  over  $\sigma$  and~$W$ is a finite set of weak constraints over $\sigma$.
Let $\cP=(P,W)$ be an optimization program over vocabulary $\sigma$ (intuitively, $P$ and~$W$ forms  {\em hard} and {\em soft} fragments, respectively). 
Set~$X$ of atoms over $\sigma$ is an {\em answer set} of $\cP$ when it is an answer set of $P$.
\end{definition}
By $\level{\cP}$ we denote the set of all levels associated with optimization program $\cP$ constructed as 
$
\{\ell \mid\, D[w@\ell,t_1,\dots,t_n]\in W\}$.
Given an answer set~$X$ of o-program $\cP$, we map $X$ and $\Pi$ to a set of tuples as follows:
$$
\weak{\cP}{X}=\{(w@\ell,t_1,\dots,t_n)\mid  D[w@\ell,t_1,\dots,t_n]\in W\hbox{ and }X\models D\};
$$
We are now ready to define a number associated with  o-program, its answer set, and a level $\ell\in\level{\cP}$:
$$
\pasw{\cP}{\ell}{X}=  \sum_{D[w@\ell,t_1,\dots,t_n]\in \weak{\cP}{X}}{w}
$$
\begin{definition}[Optimal answer sets]\label{def:oascc3}
Let $X$ and $X'$ be answer sets of $\cP$.
Answer set $X$ is {\em dominated by}  $X'$ if 
there is some integer $\ell\in \level{\cP}$ such that 
\beq
\pasw{\cP}{\ell}{X'}<\pasw{\cP}{\ell}{X}
\eeq{eq:def1} 
and
\beq
\pasw{\cP}{\ell'}{X'}=\pasw{\cP}{\ell'}{X}
\eeq{eq:def2} 
for all integers $\ell'>\ell$.

An answer set $X^*$ of $\cP$ is {\em optimal}  if there is no answer set~$X'$ of $\cP$ such that $X^*$ is dominated by $X'$.
\end{definition}



\begin{example}\label{example:ocas}
We now exemplify the definitions of an optimization program and an optimal answer set.  Recall the CAS program constructed in Example~\ref{example:cas}. Let us denote it as that $P_1$.
An optimal answer set of o-program
\beq
(P_1,\{\wr a.~ [-1@1]\})
\eeq{eq:sampleop2}
is $\{a~b\}$; whereas 
program
\beq
(P_1,\{\wr a.~ [1@1]\})
\eeq{eq:sampleop}
has two optimal answer sets $\{c\}$ and $\{c, |x\geq12|\}$. 
An optimal answer set of another o-program
$$
\begin{array}{cl}
  (P_1,   &  \{~\wr a.~ [-1@1]\\
     & ~~~\wr |x=12|.~ [-2@1]\})
\end{array}
$$
is $\{c,~~|x=12|\}$.

Let us now consider slightly more complex o-programs. 
Let $W_1$ denote the set consisting of the following weak constraints:
$$
\begin{array}{ll}
\wr a. &[-1@1]\\
       \wr b. &[-1@1] \\
       \wr a,b.& [-1@1]\\
       \wr c. &[-2@1]
\end{array}
$$
O-program  $(P_1,W_1)$ has two optimal answer sets, namely,
 \beq
\{c\}\hbox{ and }\{c, |x\geq12|\}.
 \eeq{eq:as2}   
Let $W_2$ denote the set consisting of the following weak constraints:
$$
\begin{array}{ll}
\wr a. &[-1@1,l]\\
       \wr b. &[-1@1,m] \\
       \wr a,b.& [-1@1,n]\\
       \wr c. &[-2@1,o]
\end{array}
$$
O-program $(P_1,W_2)$
 has a  unique optimal answer set
 \beq
 \{a,~b\}.
 \eeq{eq:as1}
\end{example}

It is worth noting that an alternative syntax is frequently used by answer set programming practitioners when they expresses optimization criteria:
\beq
\#minimize\{w_1@\ell_1, t_{11},\dots, t_{1{k_1}} :lits_1;~~\dots;~~w_n@\ell_n,t_{n1},\dots, t_{n{k_n}}:lits_n \},
\eeq{minimize_statement}
where 
 $k_1,\dots,k_n\geq 0$ and
$lits_i$ is of the form 
$a_1,\dotsc, a_j,\ not\  a_{j+1},\dotsc,\ not\  a_m$
so that $m>0$ and  $a_1,\dots,a_m$ are atoms.
This statement stands for $n$ weak constraints
$$
\begin{array}{l}
\wr lits_1[w_1@\ell_1t_{11},\dots, t_{1{k_1}}]\\~~\dots~~\\\wr lits_n[w_n@\ell_n,t_{n1},\dots, t_{n{k_n}}].
\end{array}
$$
Similarly, statement
\beq
\#maximize\{w_1@\ell_1, t_{11},\dots, t_{1{k_1}} :lits_1;~~\dots;~~w_n@\ell_n,t_{n1},\dots, t_{n{k_n}}:lits_n \}
\eeq{maximize_statement}
stands for $n$ weak constraints
$$
\begin{array}{l}
\wr lits_1[-w_1@\ell_1t_{11},\dots, t_{1{k_1}}]\\~~\dots~~\\\wr lits_n[- w_n@\ell_n,t_{n1},\dots, t_{n{k_n}}].
\end{array}
$$

\begin{example}\label{example:ocas1}
Consider o-program~\eqref{eq:sampleop2}. The optimization requirement
$$\wr a.~ [-1@1]$$
of that program can be stated either as
$$
\#minimize\{-1@1:a\}
$$
or as
$$
\#maximize\{1@1:a\}
$$
Set $W_2$ of weak constraints from Example~\ref{example:ocas} can be represented as
$$
\#minimize\{-1@1,l:a;-1@1,m:b; -1@1,n:a,b; -2@1,o:c\}.
$$
\end{example}

\subsection{\ezsmt3 implementation details}
We now turn our attention to the question of how the support for optimization statements is implemented within \ezsmtv3. 

We used propositional programs with weak constraints to introduce their semantics. 
Yet, \ezsmtv3 supports weak constraints with ASP variables. As for any other language constructs, \ezsmtv3 starts its processing by invoking grounder \gringo to produce a propositional program. It is due to note that \gringo makes additional transformations to weak constraints so that the resulting set of weak constraints has a simpler form than discussed in the earlier section. During this transformation auxiliary atoms are introduced into the program. When answer sets are computed for this new program the auxiliary atoms can be safely dropped to obtain the answer sets of the original program.
The weak constraints that system \ezsmtv3 is exposed to beyond the point of grounding has one of the following forms
\begin{align}
\wr\  a.~[w@\ell,t_1,\dots,t_n]\label{onewc}\\
\wr\ not\ a.~[w@\ell,t_1,\dots,t_n]\label{twowc}
\end{align}
where $a$ is an atom. 
In addition, any weak constraint that appears within a ground program produced by \gringo is such that the expression $w@\ell,t_1,\dots,t_n$ appearing in that weak constraint is unique (for example, it could be used as an identifier for this constraint in the program). Let us call a program satisfying stated conditions -- {\em gringo o-program}.

Here we avoid describing formally the procedure implemented within \gringo for ``normalizing'' optimization statements. Yet, we use our sample sets $W_1$ and $W_2$ of weak constrains from Example~\ref{example:ocas} to hint at its details.
Set $W_1$  will be rewritten by \gringo in the following style
$$
\begin{array}{ll}
aux_1\ar a.&\\
aux_1\ar b.&\\
aux_1\ar a,b.&\\
\wr aux_1. &[-1@1] \\
\wr c. &[-2@1]
\end{array}
$$
whereas set $W_2$  will be rewritten by \gringo as
$$
\begin{array}{ll}
aux_2\ar a,b.&\\
\wr a. &[-1@1,l]\\
       \wr b. &[-1@1,m] \\
       \wr aux_2.& [-1@1,n]\\
       \wr c. &[-2@1,o]
\end{array}
$$
so that $aux_1$ and $aux_2$ are some fresh auxiliary atoms.

In order to process o-programs with weak constraints of the form~\eqref{onewc} and~\eqref{twowc}, \ezsmtv3 relies on the transformations proposed by~\cite{lie23a,lie24a} in the scope of so called w-systems. W-systems are meant as an abstraction to encapsulate various logic-based formalisms extended  with optimization expressions. 
It is due to note that the semantic characterization of optimization statements utilized by~\cite{lie23a,lie24a} 
is in the tradition  stemming from partial weighted MaxSat~\citep{fu06}. We restate their semantics for the case of optimization programs studied here and point at the differences. Yet, for the case of gringo o-programs the semantics as stated here and the one studied by~\cite{lie23a,lie24a} coincide. Thus, the transformations that we mentioned in the beginning of the paragraph can be safely applied.

Let us define another number
associated with  o-program~$\cP$, its answer set~$X$, and a level $\ell\in\level{\cP}$:
$$
\pasw{\cP{^\#}}{\ell}{X}=  \sum_{D[w@\ell,t_1,\dots,t_n]\in 
 W\hbox{ and }X\models D}{w}
$$
We define a concept of pw-dominance and pw-optimal answer sets as in Definition~\ref{def:oascc3} by replacing $\cP$ with $\cP^{\#}$ in equations~\eqref{eq:def1} and~\eqref{eq:def2}.

\begin{example}\label{ex:oaspcore2}
Let us now  illustrate the difference between 
optimal and pw-optimal answer sets. 
Consider the CAS program denoted as $(P_1,W_1)$ in Example~\ref{example:ocas}. 
Its two optimal answer sets are listed in~\eqref{eq:as2}.  Its
unique pw-optimal answer set is presented in~\eqref{eq:as1}.
On the other hand, recall that~\eqref{eq:as1} is the unique
 optimal  answer set of o-program $(P_1,W_2)$. The same set forms the unique 
  pw-optimal answer sets of  $(P_1,W_2)$.
\end{example}
In the last example, when we consider o-program $(P_1,W_2)$,
 it is not by chance that its optimal and pw-optimal answer sets coincide. 
 This is a consequence of a general fact captured by the following proposition.
 \begin{proposition}\label{prop:opt}
     For o-program $(P,W)$, if  
the cardinality of a set 
\beq \{ (w@\ell,t_1,\dots,t_n)  \mid
     D[w@\ell,t_1,\dots,t_n] \in W\}\eeq{eq:setcn}
    is equal to  the cardinality of $W$, then the optimal and pw-optimal answer sets of 
    $(P,W)$ coincide.
 \end{proposition}
 Note how given  sets $W_1$ and $W_2$ from Example~\ref{ex:oaspcore2}, the sets corresponding to~\eqref{eq:setcn} follow, respectively:
$$
\begin{array}{llll}
\{\-1@1,&-2@1\}& &\\
\{\-1@1,l&-1@1,m&-1@1,n&-2@1,0\}\\
\end{array}
$$
It is easy to see that any gringo o-program satisfies the if-condition of Proposition~\ref{prop:opt}.

Now that we established that transformations studied by~\cite{lie24a} are safe for gringo o-programs we present some details on these transformations. 
First,  the weak constraints are normalized so that they only contain positive weights. Second, the weights of the weak constraints are rescaled based on the factor computed for each level while taking into account the weights of smaller levels. 
As a result, newly composed weak constraints can be considered of the same level. The first rewriting is simple. It starts by dropping all weak constraints with $0$ weight. Then, any weak constraint of the form~\eqref{onewc} that has a 
negative weight $w<0$  is replaced by the following weak constraint:
$$
\wr\  not\ a.~[- w@\ell,t_1,\dots,t_n]
$$
and any weak constraint of the form~\eqref{twowc} that has a 
negative weight $w$  is replaced by:
$$
\wr\  a.~[- w@\ell,t_1,\dots,t_n].
$$
Note how $-1 w$ results in a positive integer.
The second rewriting that eliminates all the distinct levels in favor of single level $1$ is more involved and we refer the reader to Section~5.2 by~\cite{lie24a} for the details on the procedure.
\cite{lie23a,lie24a} illustrate that the described rewritings preserve the pw-optimal models of the program. System \ezsmtv3 implements these rewritings. 

Upon the completion of the rewriting process, the \ezsmtv3 deals with the collection of weight constraints of the following form
$$
\begin{array}{l}
\wr\  a_1.~[w_1@1,t_{11},\dots,t_{1n_1}]\\
\dots\\
\wr\  a_k.~[w_k@1,t_{k1},\dots,t_{k{n_k}}]\\
\wr\ not\   a_{k+1}.~[w_{k+1}@1,t_{{k+1}1},\dots,t_{{k+1}n_{k+1}}]\\
\dots\\
\wr\ not\   a_{k+m}.~[w_{k+m}@1,t_{{k+m}1},\dots,t_{{k+m}{n_{k+m}}}]\\
\end{array}
$$
so that all weak constraints are of the same level $1$ and all weights $w_1,\dots,w_{k},w_{k+1},\dots,w_{k+m}$ are positive numbers. Given the above collection of the weak constraints, \ezsmtv3 composes the following expression in the language of the SMT-LIB:
$$
\begin{array}{ll}
(assert\ (=\ val\  
            (+& (ite\ a_1\ w_1\ 0)\ \\
            &\dots\\
             &  (ite\ a_k\ w_k\ 0)\ \\
 &              (ite\ (not\ a_{k+1})\ w_{k+1}\ 0)\ \\
  &             \dots\\
  &             (ite\ (not\ a_{k+m})\ w_{k+m}\ 0)\             \\
   &            )))
   \end{array}
$$
where variable $val$ is declared as an integer and expression {\em ite} is intuitively evaluated as an  if-then-else statement. Note how the introduction of integer variable $val$ translates into the 
use of SMT(LIA) or SMT(LIRA) logics when the SMT solver is invoked as depicted in Figure~\ref{fig:castosmt2}.

It is now due to describe the iterative procedure utilized to compute optimal answer sets.
When the first answer set of a given program with weak constraints is computed, the answer is inspected to collect the value $v$ of  variable $val$. Then the new SMT-LIB statement is composed
\beq
(assert\ (<\ val\ v))  
\eeq{eq:opt}
and the SMT solver of choice is instructed to continue its search with this new statement. The process of inspecting for the value of variable $val$ and appending the statement of the form~\eqref{eq:opt} is repeated till we establish that the problem becomes unsatisfiable. 
System \ezsmtv3 implements an anytime approach for computing optimal answer sets. In other words, it displays each found answer set to a user with the guarantee that each following answer set dominates the one presented earlier.

\begin{figure}[th]
\begin{tabular}{l|c}
    \hline\hline
    CAS(LIA)&{SMT(LIA)}\\
    \hline
    CAS(LRA)&{~~~~~~~SMT(LIRA)~~~~~~~}\\
    \hline
  CAS(LIRA)~~~~~~~&{SMT(LIRA)}\\
  \hline
    CAS(IDL)&{SMT(LIA)}\\
    \end{tabular}
    \caption{Mapping of logics from CAS programs with weak constraints to respective SMT formulas.\label{fig:castosmt2}}
\end{figure}


\section{Experimental Analysis}\label{sec:bench}
In this section, we present the results on 
comparing the performance of system 
\ezsmtv3  with the state-of-the-art solvers such as \clingcon~\citep{ost17}, \clingodl~\citep{jan17}, and \clingolp~\citep{jan17}. 
The unique part of this comparison is that all encodings used  were identical for all systems involved. This also explains the choice of systems to benchmark against. Cited papers above present experimental comparison of stated systems with other related technologies.

Three benchmarks, namely, Reverse Folding (RF), Incremental Scheduling (IS), and Weighted Sequence (WS), come from the Third Answer Set Programming Competition~\citep{aspcomp3}. We obtain the \hbox{\clingcon} encoding of IS from work by \cite{ost17}. 
We include a benchmark problem called Blending (BL) from work by  \cite{sar17}. We also add a modification of this benchmark called Mixed-BL, which contains variables over both integers and reals.
Three more benchmarks, namely, RoutingMin (RMin), RoutingMax (RMax), and Traveling Salesman (TS) are obtained from work by~\cite{liu12}.  
The original TS benchmark is an optimization problem, and we turn it into a decision problem. 
The original RoutingMax and RoutingMin problems are stated as CAS(LIA) programs.
It was possible to find a formulation of these problems using the CAS(IDL) language.\footnote{Such a reformulation was suggested by Max Ostrowski.} 
In addition, we created another variant of  the RoutingMax problem encoding by re-formulating one of its integer linear constraints in the original encoding as an aggregate ($\#sum$) expression.
The  Labyrinth (LB) benchmark is extended from the domain presented in the Fifth Answer Set Programming Competition~\citep{aspcomp5}.
This extension allows us to add integer linear constraints into the problem encoding.
Also, we present results on two benchmarks from work by~\cite{bal17}, namely, Car and Generator (GN). It is due to remark that all encodings from the literature were inspected and when possible augmented with additional domain restrictions for their constraint variables. This change was due to an observation that  systems such as \clingcon typically benefit from prespecified tighter domain on constraint variables.

System \ezsmtv3 and all used benchmarks are hosted at \url{https://github.com/ylierler/ezsmtv3} .

All benchmarks are run on an Ubuntu 20.04.6 LTS (64-bit) system with an Intel® Core™ i7-7700 CPU @ 3.60GHz with 31.2 GiB RAM. The resource allocated for each benchmark instance is limited to one CPU core and 4 GiB of RAM. We set a timeout of 1800 seconds for each instance. 
Systems that we use to compare the performance of variants of 
\ezsmtv3 (invoking SMT solver 
\cvcFour  v.~1.8; 
\cvcFive v.~1.0.8;
\yices  v.~2.6.4;
\zThree v.~4.8.7
) are 
\clingcon  v.~5.2.1, \clingolp v.~0.2.0 and \clingodl v.~1.5.0.
The \gringo system v.~5.4.0 is used as a grounder for \ezsmtv3.

Within all presented figures, 
all of the steps involved, including grounding and translation, are reported as part of the total solving time.
Letter~\eezsmtv3 stands for \ezsmtv3;
{ \eclingcon}  stand for \clingcon; and
{  \eclingodl} stands for \clingodl . The number in parenthesis after the name of the benchmark specifies how many instances were used in experiments.
The time reported is the cumulative time of all the instances of the particular benchmark. The number of unsolved instances due to timeout or insufficient memory is put inside parentheses. The cumulative time mentioned in bold font is the least time taken for that particular benchmark problem using the corresponding solvers. The ``-'' symbol is used to show that the considered solver does not support this particular encoding. 
The benchmarks are divided into categories. The acronyms T and NT in the category names indicate that the programs are tight and non-tight, respectively. 
The second part of the category name indicates the logic used to formulate the CAS encoding of the considered problems.
For non-tight (NT) programs, more solving options are possible, such as the use of different level ranking formulas using flags.
  \cite{shen18ezsmt} highlights the impact of the level ranking flags on the performance of \ezsmtPlus. Similar impact is expected on \ezsmtv3.

  Before presenting individual results let us mention that 
\ezsmtv3 can be seen as a more versatile system than its mentioned peers 
  \clingcon, \clingodl, and \clingolp. Indeed,  \ezsmtv3 supports programs of four kinds, namely, CAS(LIA), CAS(IDL), CAS(LRA), and CAS(LIRA). Systems \clingcon, \clingodl, and \clingolp support CAS(LIA), CAS(IDL), CAS(LRA) programs, {\em respectively}. This fact explains why figures that follow contain benchmark data for different subsets of systems.
  
We start the discussion of the experimental analysis with the presentation of  Figure~\ref{table3}. 
This figure is meant to illustrate the uniqueness of the \ezsmtv3 system. Unlike its other peer systems geared to support a specific logic,  \ezsmtv3  implements  various logics including LIRA. 
Thus, \ezsmtv3 is capable of solving new kinds of domains. 
In the future, we envision extensions of the system to more logics provided by the SMT-solving portfolio. 
Another special feature of \ezsmtv3 is that it can be seen as a multitude of systems. Indeed, each SMT solver invoked by the system provides us with different computational capabilities. Figure~\ref{table3} illustrates that SMT solvers \cvcFour and \cvcFive are superior to \zThree for the case of the considered benchmark. The same figure does not present timings for \ezsmtv3 invoking \yices. This is due to the fact that SMT solver \yices provides no support for LIRA logic.

\begin{figure*}[h!]
\resizebox{\textwidth}{!}{%
\begin{tabular}{c|c|rrrr}
  \hline
Category & {Benchmark}  & \eezsmtv3(\zThree) & \eezsmtv3(\yices) & \eezsmtv3(\cvcFour) & \eezsmtv3(\cvcFive) \\
\hline
\multirow{1}{*}{T-LIRA} & Mixed-BL (30)&  2957.33 & - & \textbf{93.30} & 140.66\\
\hline
\hline
\end{tabular}}
\vspace{-1em}
\caption{{Summary of Experimental Data on CAS(LIRA) Encodings}}\label{table3}	
\end{figure*}

Figure~\ref{table2} presents the comparison between \clingolp and \ezsmtv3 on CAS(LRA) encodings available. 
Figure~\ref{table5} presents the comparison between \clingcon and \ezsmtv3 on CAS(LIA) encodings. It is due to note that system \clingcon is a mature tool that has been under development for close to a decade, whereas  \clingolp has been developed to illustrate the versatility of \clingo series 5 that provides capabilities to bootstrap nontrivial extensions. 
These figures seem to indicate that relying on SMT solvers as a backend is a viable approach. We see how \ezsmtv3 variants are competitive or superior with respect to \clingolp. At the same time it is obvious that when a technology is specifically geared towards solving CAS(LIA) programs such as \clingcon, then the efforts are paid off. On several of the benchmarks, \ezsmtv3 is comparable in its performance with \clingcon, but often enough \clingcon exhibits superior performance. This consistent difference may also be due to the fact that solving technology of SMT solvers works under no assumption of finite domains such as imposed by the technology behind \clingcon.
Yet, when it is important to lift finite domain restrictions \ezsmtv3 is now an option.
Worth noting, \ezsmtv3 seems to weather a competition better in a realm of tight programs versus nontight.

\begin{figure*}[h!]
\resizebox{\textwidth}{!}{%
\begin{tabular}{c|c|r|rrrr}
  \hline
Category & {Benchmark} & \clingolp & \eezsmtv3(\zThree) & \eezsmtv3(\yices) & \eezsmtv3(\cvcFour) & \eezsmtv3(\cvcFive) \\
\hline
\multirow{3}{*}{T-LRA} & BL (30)& (2) 11640.43 & 62.39 & \textbf{37.81} & 43.52 & 54.74\\
\hhline{~------}
 & GN (8)& 4.19 & 4.23 & \textbf{4.12} & 5.04 & 4.78\\
\hline
\hline
\end{tabular}}
\vspace{-1em}
\caption{{Summary of Experimental Data on CAS(LRA) Encodings}}\label{table2}	
\end{figure*}

\begin{figure*}[h!]
\resizebox{\textwidth}{!}{%
\begin{tabular}{c|c|r|rrrr}
  \hline
Category & {Benchmark} & \eclingcon & { \eezsmtv3(\zThree)} & 
{ \eezsmtv3(\yices)} &
{ \eezsmtv3(\cvcFour)} & { \eezsmtv3(\cvcFive)} \\
\hline
\multirow{4}{*}{{ NT-LIA}}  & RMin & \textbf{1.17} & 95.81 & 91.88 & 110.67 & 107.61\\
 &(100)&&&\\
\hhline{~------}
& RMax(\#sum)& \textbf{20.34} & 35166.79 & 10802.86 & 16754.00 & 10707.14\\
 &(100)&&&\\
\hhline{~------}
& RMax(\&sum)&  \textbf{10.62} & 2087.30 & 592.44 & (100)   & 1743.65\\
 &(100)&&&\\
\hhline{~------}
 & TS (30) & \textbf{45.50} & (22) 43763.86 & (1) 2117.40 & (1) 3129.99 & 3035.31\\
 \hhline{~------}
 & LB (22) & \textbf{(1) 4445.42} & (1)\ \ \ 7648.74 &(1) 7309.14 &(1) 7397.58 &(2)  8361.58\\
\hline
\hline
\multirow{3}{*}{T-LIA} &  RF (50) & \textbf{105.63} &(1) 8417.93 &(1) 7610.42 &(20) 48819.14 &(1) 11852.58\\
\hhline{~------}
 & IS (30)& \textbf{(5) 9060.70} &(5) 9150.36 & (5) 9212.70&(5) 10074.78 &(5)\ \ \ 9331.48\\
 \hhline{~------}
 & WS (30) & \textbf{17.22} & 57.55 & 45.51 & 57.3 & 57.57\\
\hline
\hline
\end{tabular}}
\vspace{-1em}
\caption{Summary of Experimental Data on CAS(LIA) Encodings}\label{table5}	
\end{figure*}


Last but not least we present Figure~\ref{table4} that summarizes the  results for CAS(IDL) encodings. In the same table, we add lines from the earlier figure that showcase the results for the same problems encoded as CAS(LIA) programs and solved by different technologies. This table  points at the possibility to improve \ezsmtv3 by exploring other translations of aggregate expressions ($\#sum$, in this case) than these currently implemented within \ezsmtv3 (these routines the system inherits from answer set solver {\sc cmodels}~\citep{giu06}). The experimental data points at the superiority of specialized propagators for processing aggregate expressions.

\begin{figure*}[h!]
\resizebox{\textwidth}{!}{%
\begin{tabular}{c|c|r|r|rrrr}
  \hline
Category & 
{Benchmark} & 
{ \eclingcon} & 
{  \eclingodl} & 
{ \eezsmtv3(\zThree)} & 
{  \eezsmtv3(\yices)} & 
{ \eezsmtv3(\cvcFour)} &
{ \eezsmtv3(\cvcFive)} \\
& (100) &&&&&&\\
\hline
\multirow{2}{*}{NT-LIA} & 
 RMax(\#sum) & \textbf{20.34} & - & 35166.79 & 10802.86 & 16754.00 & 10707.14\\

\hhline{~-------}
&RMax(\&sum) & 
\textbf{10.62} &
- &
2087.30 & 
592.44 & 
 (100) & 
1743.65\\

 \hline
\multirow{1}{*}{{ NT-IDL}}  & RMax DL & - & \textbf{4.38} & 22682.92 & 9687.78 & 10620.85 & 16277.06 \\
\hline
\hline
\multirow{1}{*}{NT-LIA} & RMin & \textbf{1.17} & - & 95.81 & 91.88 & 110.67 & 107.61\\
 \hline
\multirow{1}{*}{{ NT-IDL}}  & RMin DL & - & \textbf{1.24} & 110.75 & 100.94 & 122.15 & 115.02\\
\hline
\hline
\end{tabular}}
\vspace{-1em}
\caption{Summary of Experimental Data on Variants of Routing Problems: CAS(IDL) and CAS(LIA) encodings combined}\label{table4}
\end{figure*}

\section{Conclusions}
This paper gives a detailed account of the \ezsmtv3 system.
A central focus of our work was the development of a robust and extensible CASP software framework 
  which may  significantly advance declarative programming and knowledge representation by offering both enhanced modeling expressiveness and access to cutting-edge solver performance.
We aimed to emulate and expand upon the success of extensible platforms such as the \clingov5 series~\citep{multishot19} and the influential SAT solver {\sc minisat}~\citep{een03}, both of which served as blueprints for designing modular, API-driven solvers. Also, the extensibility of {\sc minisat} led to more than a decade of impactful developments in SAT and related technologies, including its use in solvers like {\sc cmodels}~\citep{giu06} and {\sc minisat(ID)}~\citep{CatBBD14}. Similarly, the flexibility of \clingov5 enabled the rapid prototyping of new CASP solvers such as \clingolp and \clingodl~\citep{jan17}.

To validate our claim that the \ezsmtv3 system is capable to support the rapid development of new CASP technologies we  bootstrap four distinct CASP solvers, one that support linear integer constraints, another one that supports constraints over reals, then one that supports mixed real integer constraints and difference logic constraints. All of these were implemented using the same streamlined methodology 
that we carefully document here. One of the intentions of this description is to attract  broader community involvement with the  \ezsmtv3 framework with the potential of seeing new solvers that support other logics widely used within SMT. 
We focused on developing a clear and accessible interface for expressing and reasoning with different kinds of constraints. This required the introduction of new language features, and streamlined integration with SMT solver technology via an  incremental solving interface. The experiment section articulates the validity of the approach and properly places the system among its peers.

One more observation is due. The \ezsmtv3 system can be seen as an alternative to SMT-LIB front-end to SMT solvers. As such, it provides a declarative programming language based on logic programming conventions to this automated reasoning technology. A similar idea was explored by the {\sc ezscp} system~\citep{lierbal17} in the scope of constraint satisfaction processing. 
That system utilized CSP solvers to process CAS programs. An alternative view to that work was
simplifying utilization of CSP solvers by providing them with the convenient interface through declarative programming languages based on logic programming. 

As a direction for future work it is important to explore the possibility of incorporating various features
supported by distinguished CASP systems. For example, systems \clingolp and \clingodl allow a programmer to use irregular atoms in  heads of the rules. System \clingcon  supports additional constraint types, such as {\em all-different}, and permits stating optimization conditions over linear sums of values of constraint variables. This direction requires deeper theoretical and practical investigation on how to incorporate such features and capabilities into \ezsmtv3.

\paragraph{Acknowledgments} We are grateful to Nicholas Wilson for his contributions to the original prototype of \ezsmtv3 as part of his Master Project (thesis equivalent) at the University of Nebraska Omaha.
We are thankful 
to Zachary Hansen for his assistance to build a sand-basket for \ezsmtv3 available on the   University of Nebraska Omaha server and providing his valuable remarks on the complete draft of the paper. Last but not least we are grateful to anonymous reviewers for taking their time to provide detailed comments on the paper. 


\paragraph{Competing interests} The author(s) declare none.

\bibliography{shared}
\end{document}